\documentclass[10pt,twocolumn,letterpaper]{article}

\usepackage{iccv}
\usepackage{times}
\usepackage{epsfig}
\usepackage{graphicx}
\usepackage{amsmath}
\usepackage{amssymb}
\usepackage{caption}
\usepackage{multirow}
\usepackage[page,header]{appendix}
\usepackage{titletoc}

\usepackage[pagebackref=true,breaklinks=true,letterpaper=true,colorlinks,bookmarks=false]{hyperref}

\iccvfinalcopy % *** Uncomment this line for the final submission

\def\iccvPaperID{****} % *** Enter the ICCV Paper ID here

\ificcvfinal\fi

\begin{document}

%%%%%%%%% TITLE - PLEASE UPDATE
\title{One-Shot Stylization for Full-Body Human Images}

\author{Aiyu Cui  \qquad  Svetlana Lazebnik \\
University of Illinois Urbana-Champaign \\
{\tt\small \{aiyucui2,slazebni\}@illinois.edu } \\ 
}

\twocolumn[{%
\renewcommand\twocolumn[1][]{#1}%
\maketitle
\begin{center}
\vspace{-8mm}
    \centering
    \includegraphics[width=0.85\textwidth]{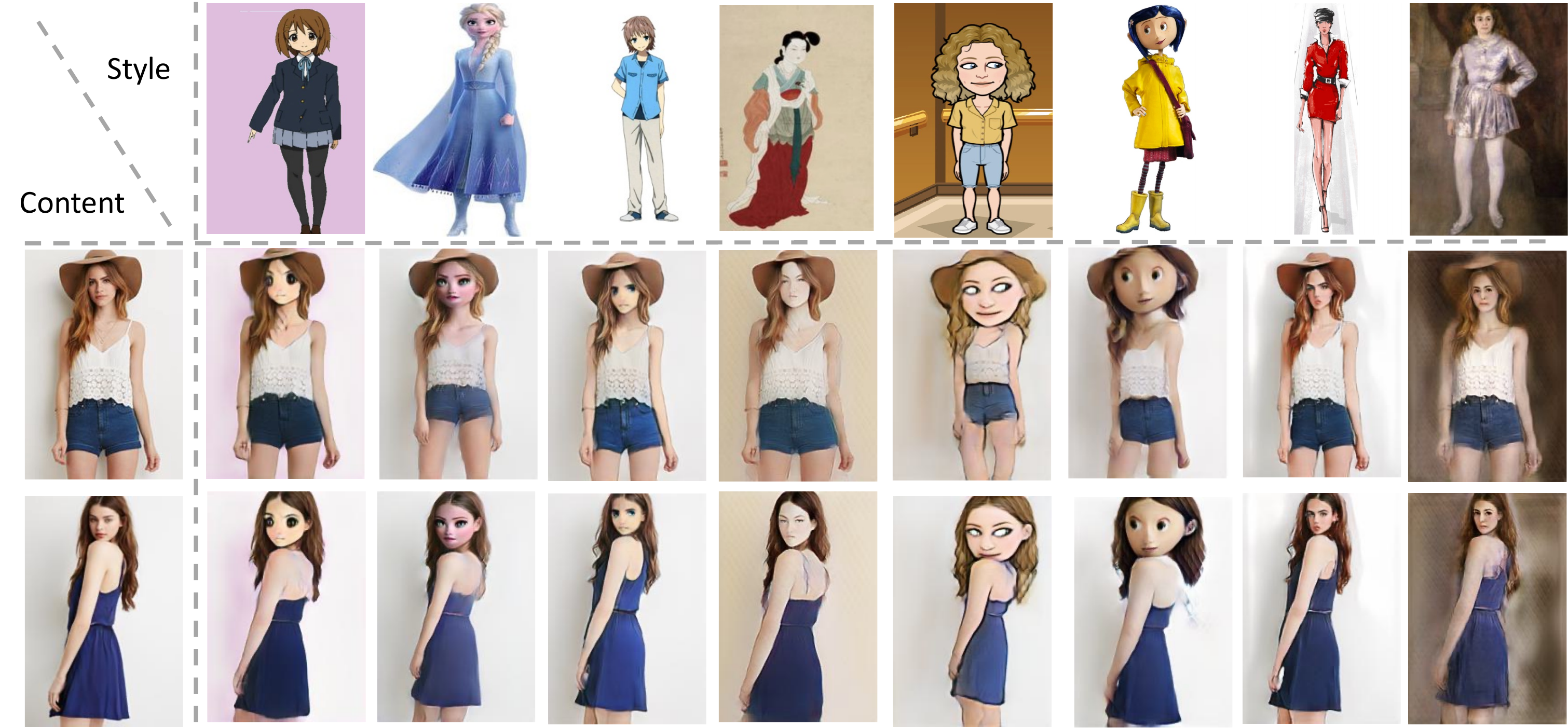}
    \vspace{-2mm}
    \captionof{figure}{This work performs one-shot stylization for full-body human images, which is the task of modifying a human photo (left column) to conform to the style of specific artistic characters (top row). 
    }
    \label{fig:teaser}
    
    \vspace{-2mm}
\end{center}%
}]

\maketitle
\begin{abstract}
    The goal of human stylization is to transfer full-body human photos to a style specified by a single art character reference image. 
    Although previous work has succeeded in example-based stylization of faces and generic scenes, full-body human stylization is a more complex domain. This work addresses several unique challenges of stylizing full-body human images. We propose a method for one-shot fine-tuning of a pose-guided human generator to preserve the ``content'' (garments, face, hair, pose) of the input photo and the ``style'' of the artistic reference. Since body shape deformation is an essential component of an art character's style, we incorporate a novel skeleton deformation module to reshape the pose of the input person and modify the DiOr \cite{Cui_2021_dior} pose-guided person generator to be more robust to the rescaled poses falling outside the distribution of the realistic poses the generator is originally trained on. Several human studies verify the effectiveness of our approach. 
\end{abstract}

%Unlike faces, for which dominant generation architectures like StyleGAN~\cite{karras2020stylegan2} exist, there is no definitive pre-trained model for full-body human generation. Thus, we investigate existing full-body generation models and 

\section{Introduction}
Image stylization has become popular in our daily life with the advent of camera filters that can render our faces using cartoon styles. 
%Such a stylization is often required to learn from a single style reference image because it is not realistic to collect a large set of a specific art style.
This work aims to learn a stylization model that can transform any full-body human photo into the style of a single art image. We address two unique challenges of the full-body person domain: 1) fine-tuning a person generator to match a given style while maintaining the original person's attributes like clothes, hair, face, and pose; and 2) rescaling an input human skeleton to better conform to the proportions of the artistic reference.

\begin{figure*}
  \centering
  \includegraphics[width=0.9\textwidth]{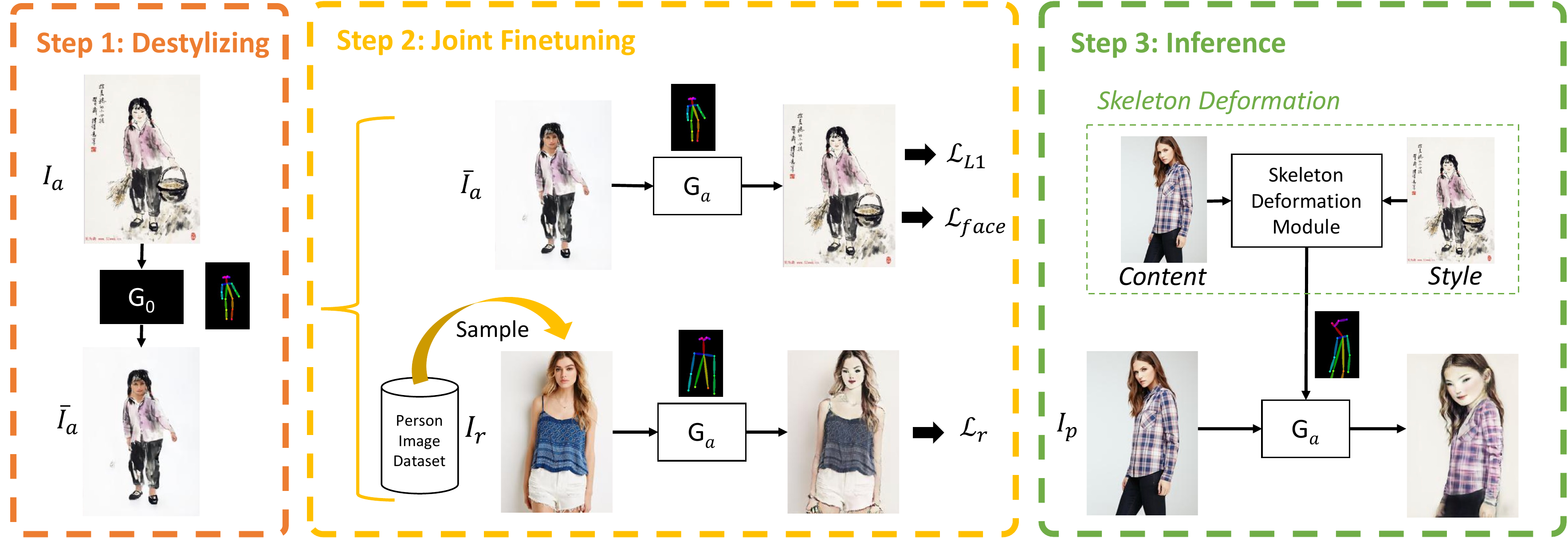}
  \vspace{-3mm}
  \caption{\textbf{Overview.} Step 1: Given an art reference  $I_a$, we ``destylize'' or reconstruct it in the photo domain by extracting the conditioning information (pose, appearance) from the reference and using it as input to a conditional person generator $G_0$ pre-trained on photos. This generator reconstructs the art image as $\bar{I}_a$. Step 2: We fine-tune our artistic generator, $G_a$, to recover $I_a$ from $\bar{I}_a$, while also enforcing style and content losses on the output of the generator for randomly sampled training person images $I_r$. Step 3: Given a test person image $I_p$, we use a learned skeleton deformation module to rescale the skeleton to match the proportions of the art reference $I_a$. Finally, we condition the fine-tuned generator $G_a$ on the rescaled pose to produce the stylized output.}
  \label{fig:overview}
  \vspace{-5mm}
\end{figure*}

Generic or universal image stylization techniques~\cite{li2017ust, kim2020dst, kolkin2022nnst} can produce compelling results for certain tasks, such as stylizing images of general scenes using painterly styles, by matching low-level feature distributions. However, as our comparisons will show, this is not sufficient for human body stylization. 
Instead, more domain-specific methods are necessary, ones that are aware of the structure and attributes of the human body.
Face stylization is another relatively mature area, with several techniques~\cite{chong2021jojogan, kwon2022oneshotclip, shah2022multistylegan, zhu2021mindthegap} based on fine-tuning of pre-trained unconditional face models like StyleGAN \cite{karras2020stylegan2}. %generating face from a latent space that can be manipulated for different editing tasks. 
However, a monolithic unconditional generation model is not sufficient to enable accuracy and controllability for full-body human generation due to outfit diversity and pose variance.
To be able to control pose and appearance, state-of-the-art human generators are separately conditioned on pose and appearance inputs~\cite{albahar2021posewithstyle, Cui_2021_dior, ren2020gfla, ren2022nted}; otherwise, learned pose will be entangled with appearance, causing editing difficulty. We show how to fine-tune such conditional person generators for full-body stylization. As an additional component of our method, we also learn to rescale the input pose to conform to the reference body shape. 
%Note that this conditioning makes such human generators image-to-image translators transferring an input person to a different pose or outfit, rather than random human image samplers. Therefore, face stylization methods relying on style mixing in the latent space~\cite{chong2021jojogan, shah2022multistylegan, kwon2022oneshotclip} are further inapplicable in full body human stylization scenarios.  

%One may argue that the universal style transfer methods \cite{li2017ust, kim2020dst, kolkin2022nnst}, which transfer any content to any style, can solve this full-body person problem. However, the style transfer manipulating style in low-level features is not sufficient to fool human eyes. Instead, fine-tuning a pre-trained person generator would ensure domain-specific semantic-aware style transfer. Also, most universal style transfers can only stylize the content pixel-wisely, while we require body shape deformation to fully represent a style. 

\begin{figure}
  \centering
  \includegraphics[width=0.48\textwidth]{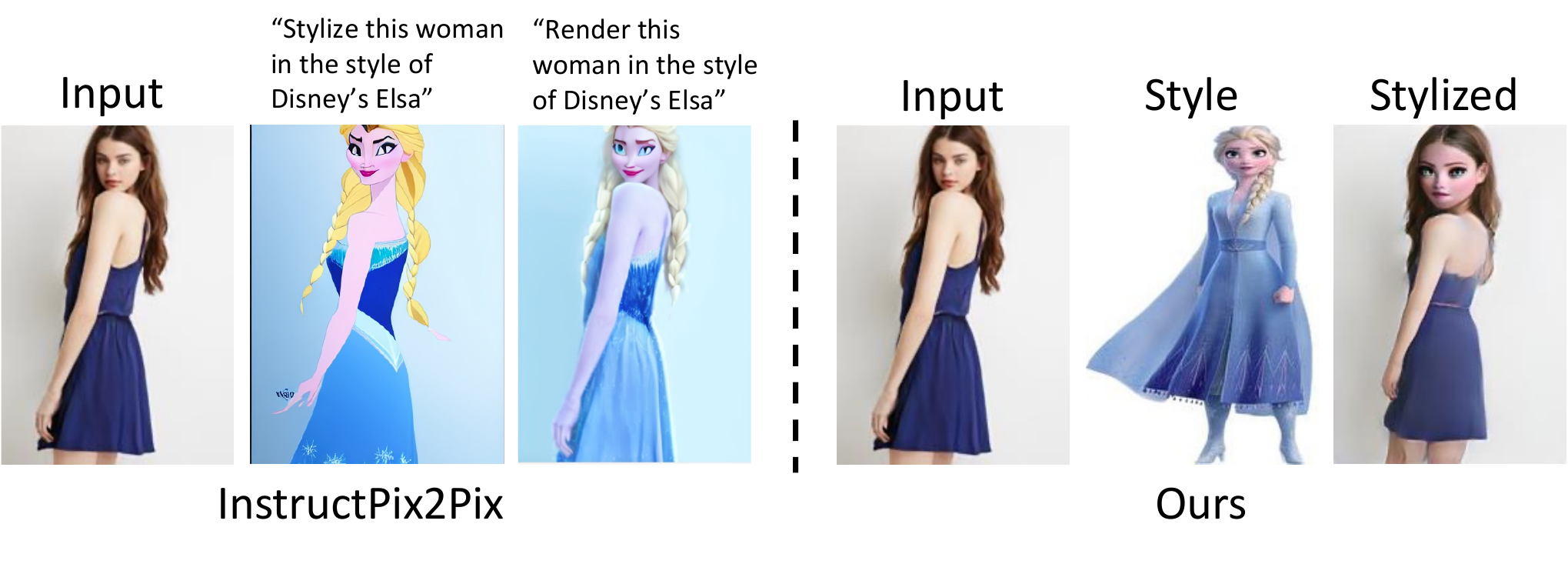}
  \vspace{-8mm}
  \caption{Stylizations produced by online demo of InstructPix2Pix~\cite{brooks2022instructpix2pix}, vs. the output of our approach.}
  \label{fig:instructpix2pix}
  \vspace{-5mm}
\end{figure}

Over the last year or two, text-guided diffusion models ~\cite{Ramesh2022Hierarchical,rombach2022stablediffusion, saharia2022imagen} have revolutionized the generation of both photo-realistic and artistic imagery. However, much of the power of these models comes from their open-endedness and "surprise element," and they do not offer the close control that example-based image stylization requires. Diffusion models are becoming increasingly adopted for image editing tasks, one recent example being InstructPix2Pix~\cite{brooks2022instructpix2pix}, which can give plausible results for text-guided person stylization, as shown in Figure \ref{fig:instructpix2pix}. However, it does not always preserve content faithfully, and cannot be used for styles that are specified using example images.

We base our method on existing conditional person generation models for tasks like pose transfer and virtual try-on. In particular, we enhance the DiOr model \cite{Cui_2021_dior} to be more robust to distorted body shapes common in cartoons, like unrealistically enlarged heads or elongated limbs. In addition, our system includes a novel skeleton deformation module to rescale the input human pose to conform to the proportions of a specified reference art image, and a fine-tuning method that can adapt any pose-guided human generation model towards a given artistic reference. An overview of our system is shown in Figure \ref{fig:overview}. Details of the system will be presented in Section 3, and ablations and user studies in Section 4 will demonstrate its promise.

\section{Related Work}
\subsection{Image Stylization}
\noindent{\bf Universal style transfer}, which uses single example images to specify style and content, has been studied for a long time, reaching new heights with deep convolutional models~\cite{chandran2021adaconv, li2017ust, kim2020dst, kolkin2022nnst}. 
These methods mainly work by transforming low-level features and do not produce satisfactory results on full-body humans, which is a narrow domain but one to which we are very perceptually attuned. Plus, most universal style transfer methods are not capable of body shape deformation. Although Deformable Style Transfer (DST)~\cite{kim2020dst} can warp the content image to match the style by predicting alignment from a set of keypoints, it is not sufficient to handle the full human body deformation, which will be confirmed by our experiments in Section 4. 

\noindent{\bf One-shot face stylization} is highly relevant to full-body stylization. Recent one-shot face stylization methods~\cite{chong2021jojogan, kwon2022oneshotclip, shah2022multistylegan, zhu2021mindthegap} succeed by fine-tuning the well-trained face generator StyleGAN~\cite{karras2020stylegan2}.
JoJoGAN~\cite{chong2021jojogan} and MultiStyleGAN~\cite{shah2022multistylegan} leverage the style mixing property of StyleGAN to obtain paired training data for a given reference style. Mind the Gap~\cite{zhu2021mindthegap} and one-shotCLIP~\cite{kwon2022oneshotclip} propose novel loss terms that take advantage of the CLIP embedding space~\cite{radford2021clip} and further uses the style code mixing property of StyleGAN to improve the style similarity. Since StyleGAN has not been directly applied to full-body human generation, those methods are not suitable for us. 

\noindent{\bf Full-body stylization.} There exist a few works that attempt full-body person stylization, like AnimeGAN~\cite{chen2020animegan} and WebtoonMe~\cite{back2022webtoonme}, but they focus on a single style -- namely, anime -- and rely on a large dataset of style images for training. They also do not perform any body deformation.

\noindent{\bf Diffusion models}~\cite{Ramesh2022Hierarchical,rombach2022stablediffusion, saharia2022imagen} hold a lot of potential for image editing. Their most straightforward use case is for text-guided inpainting~\cite{glide,meng2022sdedit}. DreamBooth~\cite{ruiz2022dreambooth} and CustomDiffusion~\cite{kumari2022customdiffusion} can take a few images of a specific object as input and associate it with a new text token, enabling the object to be rendered in various novel contexts specified by text prompts. More relevant to us, InstructPix2Pix~\cite{brooks2022instructpix2pix} can stylize a person image if the target style is specified by a text prompt. However, as can be seen from Figure \ref{fig:instructpix2pix}, prompt engineering can greatly affect the results and there are no precise control guarantees. We are interested in stylization based on a reference image, without text involvement, while requiring the generator to stay close to the pose, hair, and garments of the input person. At present, text-to-image diffusion models are not yet a good fit for this setting.

\subsection{Full-body Human Generation}
%We take the approach of fine-tuning a conditional full-body human generation model. A
Although unconditional human image generation models have recently been introduced~\cite{fu2022styleganhuman,fruhstuck2022insetgan}, conditional models~\cite{albahar2021posewithstyle,Cui_2021_dior, han2019clothflow,tang2020xinggan, ren2020gfla,zhou2022casd,zhu2019patn} are more suitable for our goal, since they provide both pose control and accurate outfit generation. These models are conditioned on an appearance input (typically segmented) and a pose skeleton, specified either by 2D keypoints~\cite{cao2019openpose} or DensePose~\cite{guler2018densepose}. The 2D keypoint representation is the most convenient for our purposes, and we do not consider DensePose. 

To accurately reproduce diverse garments, a good choice is given by warping-based pose transfer methods~\cite{Cui_2021_dior, han2019clothflow, ren2020gfla}, which first predict an explicit flow field to transform the source person to the target pose and then use a generator to render the warped person. Clothflow~\cite{han2019clothflow} is infeasible for our end-to-end fine-tuning due to its multi-stage design. GFLA~\cite{ren2020gfla} predicts flow fields to warp at feature levels before decoding it to the image space. DiOr~\cite{Cui_2021_dior} extends GFLA to allow separate control of each garment to further support virtual try-on tasks.

Another family of pose transfer models are those designed in an encoder-decoder manner~\cite{lewis2021tryongan, men2020adgan, ren2022nted,zhu2019patn,tang2020xinggan,zhou2022casd}. ADGAN~\cite{men2020adgan} and CASD~\cite{zhou2022casd} encode each body part separately and concatenate the style codes together before decoding. TryOnGAN\cite{lewis2021tryongan} is a StyleGAN architecture conditioned on pose but not input person, so it does not fit our purpose. Other works~\cite{ren2022nted, tang2020xinggan, zhu2019patn} learn to implicitly generate the input person and target pose, where NTED~\cite{ren2022nted} uses an architecture with skip connections similar to StyleGAN and achieves SOTA performance on pose transfer. We will see that these encoder-decoder models can render realistic images but suffer from reconstruction errors in decoding. 

% InsetGAN is a unconditional person generation model so not relevant

\section{Proposed Method}

Given a single reference art image $I_a$, our goal is to produce a generator $\mathbf{G}_a$ that can turn any realistic photo $I_p$ into an output $I'_p$ matching the style of $I_a$. 

Like prior one-shot face stylization models \cite{chong2021jojogan, shah2022multistylegan, kwon2022oneshotclip} that are fine-tuned from pre-trained face generators, we obtain such a generator by fine-tuning a pre-trained person generation model. Our base generator is denoted by $\mathbf{G}_0(I, \Phi)$, where $I \in \mathbb{R}^{3\times H\times W}$ is the source image for appearance and $\Phi \in \mathbb{R}^{18\times H\times W}$ is the pose represented by $18$ keypoints in OpenPose~\cite{cao2019openpose} format. The output is the source person in the target pose. 
Note that many pose-guided person generators, like DiOr, also take a segmentation input to control the outfit, but we only show the $I$ and $\Phi$ as $\mathbf{G}_0$'s inputs for simplicity.

As shown in Fig. \ref{fig:overview}, we first destylize $I_a$ to obtain a more ``realistic'' input $\bar{I}_a$, and then fine-tune $\mathbf{G}_0$ to obtain $\mathbf{G}_a$ by attempting to reproduce $\bar{I}_a$ from $I_a$. To counteract the tendency of the generator to overfit the art reference, we additionally impose stylization losses on batches of real person images. Further, during the inference step, we deform the pose $\Phi_p$ of the input person $I_p$ to $\Phi'_p$, to better align the body proportions to the reference art input, using a learnable skeleton deformation module. The final stylized output is obtained as $I'_p = \mathbf{G}_a(I_p, \Phi'_p)$.

\subsection{Base Model Selection}

\begin{figure}
  \centering
  \includegraphics[width=0.48\textwidth]{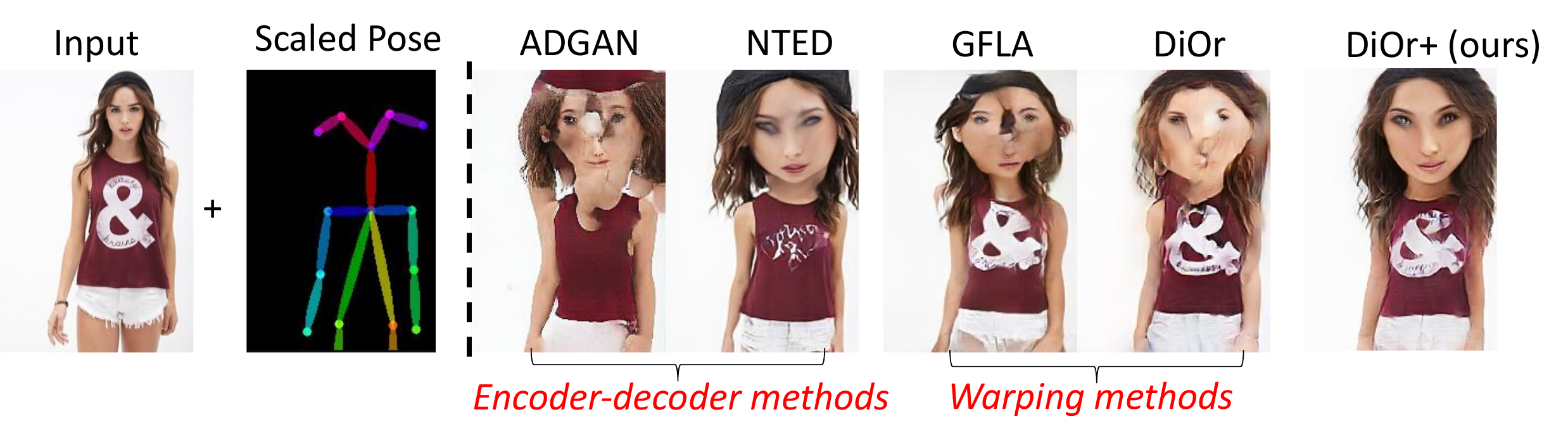}
  \vspace{-3mm}
  \caption{Illustration of generator performance when conditioned on scaled poses (in this case, an enlarged head). Encoder-decoder methods fail to reconstruct garment details, while warping-based methods suffer from the outlier proportions.}
  \label{fig:base_model_pose_transfer}
  \vspace{-5mm}
\end{figure}

Although our method is designed to work with generic pose-guided person generation models, the choice of the base model $\mathbf{G}_0$ is essential to the stylization quality. 
Unlike StyleGAN for face generation, there is no dominant model for the full-body human generation task. Encoder-decoder architectures \cite{ lewis2021tryongan, men2020adgan,ren2022nted} are robust to outlier body ratios (e.g., large faces), but suffer reconstruction errors for complex patterns when decoding from the embedding space. Warping-based methods \cite{Cui_2021_dior,ren2020gfla} are often not robust enough for extreme body deformations, as shown in Fig. \ref{fig:base_model_pose_transfer}.

The recent pose transfer model DiOr \cite{Cui_2021_dior} is suitable for our stylization task, given that it both preserves garment details well and has the flexibility to support additional editing tasks like virtual try-on. DiOr takes a source image, a skeleton, and a semantic segmentation mask as input. 
DiOr separately controls the face/skin and each garment: it first generates the human face and skin from a $64\times64$ feature map broadcast with mean skin vectors in the foreground and the mean background vector in the background. Then it recurrently layers warped garments on top of the generated skin at feature resolution $64\times64$ before decoding back to the image space. Consequently, one can perform virtual try-on with DiOr by inputting or swapping a garment with a new one during the recurrent generation.

We found that DiOr still struggles to reproduce large faces, so we make three modifications to its training. First, we change the skin encoding to a $1\times512$ vector and use AdaIN~\cite{huang2017adain} to generate the face and skin's feature maps conditioned on the input pose features at $16\times16$ resolution, rather than decoding a broadcast feature map at $64\times64$ resolution. 
Second, at training time, we randomly enlarge the input target person from $1\times$ to $2\times$ so that it better learns to handle large faces. Moreover, while DiOr is jointly trained with 80\% pose transfer and 20\% image inpainting, we increase the inpainting task ratio to 50\%, which makes the model better at preserving garment details. We call the resulting generator \textbf{DiOr+} in the following. Fig. \ref{fig:base_model_pose_transfer} shows that it suffers from fewer artifacts than other generators when conditioned on scaled poses (e.g., abnormally large faces), and experiments of the next section further confirm this. 

\subsection{Fine-tuning}
\noindent \textbf{Destylizing the reference art input.}
%We first destylize the one-shot art reference input $I_a$ to the photo domain, as $\bar{I}_a$, which will later be used to fine-tune a human generation model to a stylization model. 
Given the art reference $I_a$, we obtain the pose $\Phi_a$ and encode the appearance in the way required by each particular person generator. E.g., DiOr~\cite{Cui_2021_dior} requires a semantic segmentation map where the classes correspond to face, skin, garments, etc. Then the output $\bar{I}_a = \mathbf{G}_0(I_a, \Phi_a)$ can be thought of as a destylization or a projection of the artistic reference into the space of the images it is trained on. 
\begin{figure}
  \centering
  \includegraphics[width=0.48\textwidth]{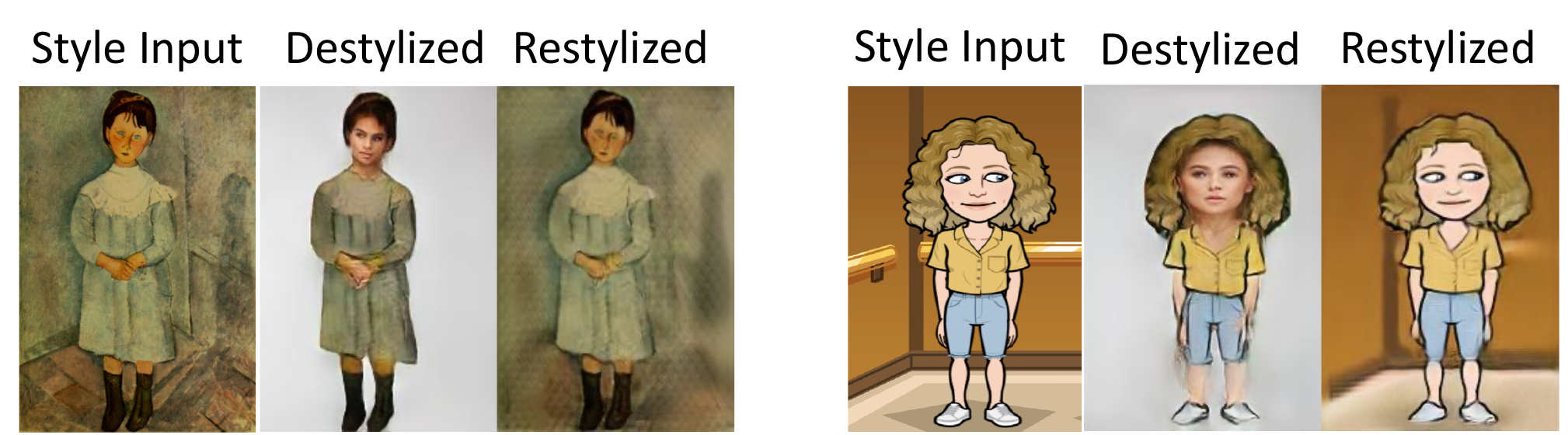}
  \vspace{-3mm}
  \caption{Destylization is accomplished by encoding all required information (pose, appearance) from the art reference and then conditioning the photo-realistic generator $\mathbf{G}_0$ on this information. Restylization is accomplished by the fine-tuned generator $\mathbf{G}_a$ with a reconstruction loss on the original art image. Note that the background is not part of the conditioning information; its reappearance in the restylized image is a sign of $\mathbf{G}_a$ overfitting to the art reference.}
  \label{fig:destylization}
  \vspace{-5mm}
\end{figure}

\noindent \textbf{Fine-tuning the human generation model.}
Initialized with $\mathbf{G}_0$, the stylization model $\mathbf{G}_a$ is fine-tuned by learning to restylize the destylized art reference image, with L1 loss for reconstruction:
\begin{equation}
    \mathcal{L}_{L_1} = ||\mathbf{G}_a(\bar{I}_a, \Phi_a) - I_a||_1.
\end{equation}
To produce a sharper and clearer face, we add a face embedding loss:
\begin{equation}
  \mathcal{L}_{face} = ||\mathbf{E}(\mathbf{G}_a(\bar{I}_a, \Phi_a)) - \mathbf{E}(I_a)||_1,
\end{equation}
where $\mathbf{E}$ is the pre-trained face encoder for inverse StyleGAN \cite{richardson2021inversestylegan}. Fig. \ref{fig:destylization} gives examples of destylization and restylization.

Since we only have a single pair $(I,\bar{I}_a)$, the above fine-tuning can easily overfit after a small number of iterations.
However, if we have access to a photographic person image dataset (we use 400 images in practice), we can regularize this fine-tuning with a joint task, which takes a randomly sampled person image $I_r$ as input and outputs its stylization $\mathbf{G}_a(I_r, \Phi_r)$ for every iteration.
We apply the Gram matrix style loss \cite{johnson2016styleloss} and the LPIPS perceptual loss \cite{zhang2018lpips} for this style transfer:
\begin{equation}
    \mathcal{L}_{r} = \mathcal{L}_{sty}(I_a, \mathbf{G}_a(I_r, \Phi_r)) + \mathcal{L}_{LPIPS}(I_r, \mathbf{G}_a(I_r, \Phi_r)).
\end{equation}

The final objective for fine-tuning is thus
\begin{equation}
    \mathcal{L} = \lambda_{L_1}\mathcal{L}_{L_1} + \lambda_{face}\mathcal{L}_{face}+ \lambda_{r}\mathcal{L}_{r},
\end{equation}
where $\lambda$ is the coefficient of each loss term.

Note that a possible alternative to training with an auxiliary set of person images $I_r$ is to fine-tune $\mathbf{G}_a$ separately for every test image, i.e., $I_r = I_p$. In this way, the model may be better able to preserve the details of a specific person, especially for images with outlier poses/garments not seen by the base model. However, for efficiency, we prefer a generator $\mathbf{G}_a$ that can be applied to any input person once it is trained, so we did not adopt this per-image optimization strategy. 

\begin{figure*}
  \centering
  \includegraphics[width=0.98\textwidth]{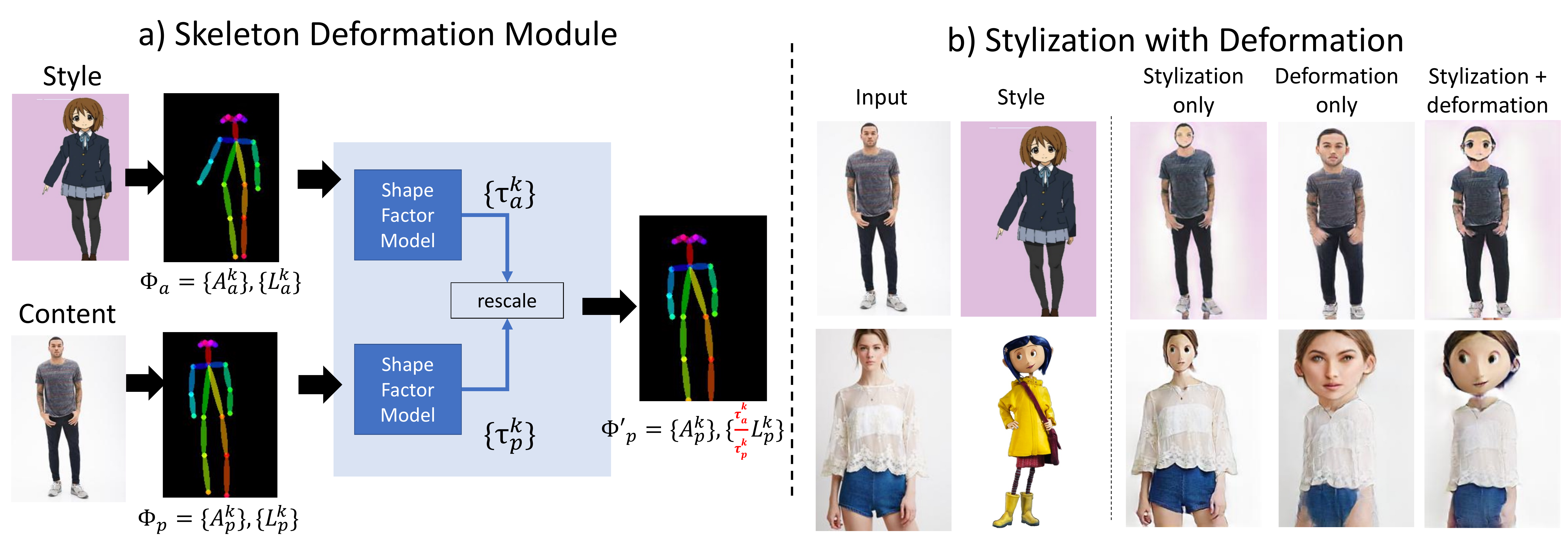}
  \vspace{-3mm}
  \caption{\textbf{Skeleton Deformation Module.} (a) Predicting body ratio factors ${\tau}$ to deform a skeleton to conform to an art style. (b) Separate effects of stylization and deformation on generator output.}
  \label{fig:deformation}
  \vspace{-5mm}
\end{figure*}

\subsection{Learnable Skeleton Deformation}

At inference time, given an arbitrary person image $I_p$ with pose $\Phi_{p}$, we can use the fine-tuned generator $\mathbf{G}_a$ to produce $\mathbf{G}_a(I_p, \Phi_{p})$. 
However, for reference images with large body deformations, it is not sufficient to stylize the input person while preserving the original pose keypoints $\Phi_p$. Therefore, we propose a learnable skeleton deformation module to rescale $\Phi_p$ to the desired art style, obtaining the modified keypoints $\Phi'_p$, before running $\mathbf{G}_a$. 

We represent a pose $\Phi$ as a connected tree with 18 nodes corresponding to body joints. Each joint has a length $l$, which is the length of the segment connecting it and its parent joint, and an angle $\alpha$ between the two segments that it connects. We use the neck joint as the root joint, whose parent joint is set as the world origin.  So, we can write 
\begin{equation} \nonumber
    \Phi = \{(\alpha_0,l_0), ..., (\alpha_{17},l_{17}) \} = (\{\alpha_i\}, \{l_i\}) = (A,L).
\end{equation}

In 3D, given a complete source human pose and a complete art pose, we can easily switch the human's body ratio without changing the gesture by changing the segment lengths while keeping the joint angles the same. 
However, direct scaling of 2D skeletons is prone to artifacts due to ignoring projection effects such as foreshortening. To mitigate such artifacts, rather than resizing each segment length individually, we group the body segments and resize them in groups. The body segments are divided into six groups: \textbf{head, shoulders, arms, torso, waist, and legs}. The left-right symmetric body parts are placed in the same group (e.g., left shoulder and right shoulder), so side poses will have their left and right body parts scale up or down concurrently to keep their projected ratio.
Now, we have the pose representations as 
\begin{equation} \nonumber
    \Phi = (A,L) = \{(A^0,...A^5), (L^0,...L^5)\} = \{A^k\}, \{L^k\}.
\end{equation}

As shown in Fig.\ref{fig:deformation}-a, Given a pose $\Phi$, we train an MLP model $\mathbf{M}$ to predict a body ratio factor $\tau$ for each group of body segments as 
\begin{equation} \nonumber
\{\tau^k\} = \mathbf{M}(\Phi).
\end{equation}

In this way, any human skeleton $\Phi_p$ can be scaled to the body ratio of any art pose $\Phi_a$'s body ratio using the respective factors $\{\tau_k^p\}$ and $\{\tau_k^a\}$ as
\begin{equation*}
    \Phi'_p = (\{A^k_p\}, \{\frac{\tau^k_a L^k_p }{\tau^k_p}\}),
\end{equation*}
where $\Phi'_p$ denotes a skeleton in person $I_p$'s pose but scaled to the body ratio of the art reference.

To train $\mathbf{M}$, we randomly scale the lengths of the skeletons extracted from a human image dataset and attempt to predict the scale factors to recover the original pose. For this, we use the L1 loss between the predicted and ground truth skeleton lengths. We also train a pose completion model to first predict missing joints for incomplete poses, which is done by reconstructing randomly masked poses via L1 loss on the Euclidean pose coordinates. 

Note that the training of this scale factor prediction model is done once up front, and is separate from the fine-tuning of the stylization model. We have tried to fine-tune the generator while conditioning on the scaled pose for $I_r$, but found the style transfer loss $\mathcal{L}_r$ to be ineffective in that case because there is no pixel-wise alignment between a scaled stylized image and the ground truth person image.

\section{Experiments} \label{sec:experiments}
\begin{figure*}
  \centering
  \includegraphics[width=0.75\textwidth]{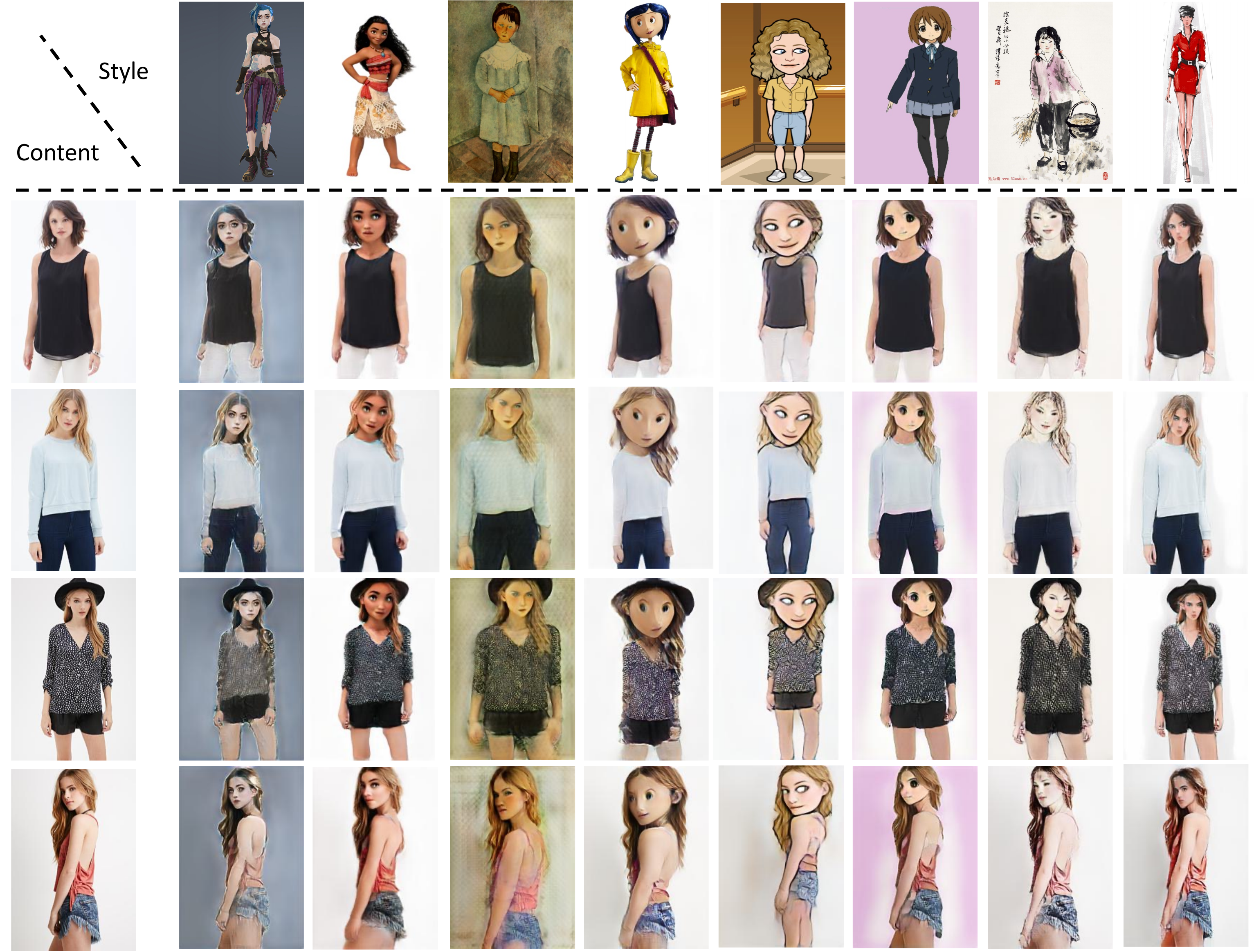}
  \vspace{-3mm}
  \caption{\textbf{Stylization results} from the full proposed methods. }
  \label{fig:example-style}
  \vspace{-5mm}
\end{figure*}

\subsection{Implementation Details}

We train the base generator on the DeepFashion dataset~\cite{liu2016deepfashion} at $256\times 176$ resolution and follow the train/test split of prior work~\cite{ren2020gfla, zhu2019patn}, %$101,966$ training pairs from $48,674$ training images and $8,570$ test pairs from $4,038$ test images, where each pair are images of the same person in the same outfit but in different poses.
consisting of $48,674$ training and $4,038$ test images. For stylization fine-tuning, we randomly sample $400$ images from the training set. 
%We train the base model for pose transfer following the same manner as prior work~\cite{Cui_2021_dior}. We fine-tune the stylization model with training set and use the test set to validate our results for both metrics and visualizations. 
We preprocess the human images by detecting 18 keypoints via OpenPose~\cite{cao2019openpose} and performing human parsing (semantic segmentation into skin, garments, etc.) via SCHP~\cite{li2020schp}. For inference, we clean the test set by removing images with no faces (filtered by whether the nose joint is detected), resulting in  $3,080$ images. 
For art references, we collect a set of 26 images from the Internet. % (see supplementary material for all of them). 
We use SCHP to perform human parsing and OpenPose for keypoint detection and manually fix any mis-detected joints. This is a feasible amount of effort because once the skeleton is obtained, we can train a stylization model to turn any person image into this style. 

%For our base person generation model, we modify DiOr\cite{Cui_2021_dior}, which separately encodes each body part, next generates the skin from a broadcasted mean skin feature map at $64\times 64$ and then layers each warped garment recurrently on the generated skin feature maps at $64\times 64$ before finally decoding back to the image space. In our modification, we change the skin encoding part by encoding the face into a $1\times512$ vector and use AdaIN~\cite{adain} to generate the skin feature maps conditional on the input pose features at $16\times16$. We train the network in the same way as DiOr's, except we randomly enlarge the input target person  from $1\times$ to $2\times$.
For the pose factor estimator, we use a 5-layer MLP with an embedding size of 256 and train it for 4,000 iterations. For fine-tuning the base model, we use a learning rate of $2e-4$ for 200 iterations with batch size 2 and set the loss coefficients to $\lambda_{rec}=200, \lambda_{face}=1, \lambda_{ST}=1$. The settings are the same for every style.
It takes about 10 minutes to fine-tune a model for a given style and less than one second to stylize each image on a single Titan V graphics card.

\begin{figure}
  \centering
  \includegraphics[width=0.42\textwidth]{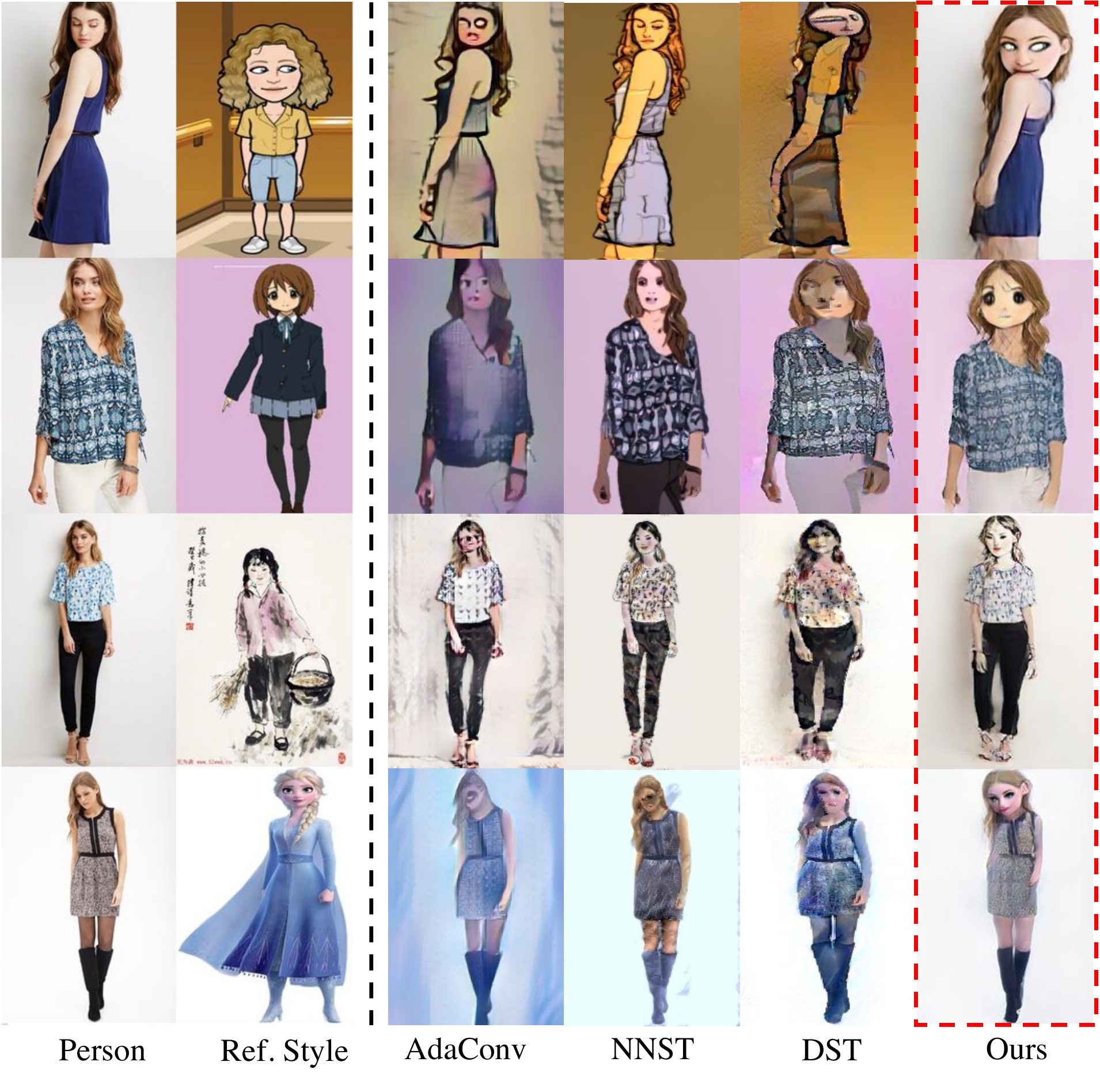}
  \vspace{-3mm}
  \caption{\textbf{Comparisons with universal style transfer.} We apply SOTA style transfer on human images with the input person as the content input and the reference art image as the style input. In the last column, we show the results of stylization from our method. }
  \vspace{-5mm}
  \label{fig:sota_comparison}
\end{figure}
\begin{figure}
  \centering
  \includegraphics[width=0.42\textwidth]{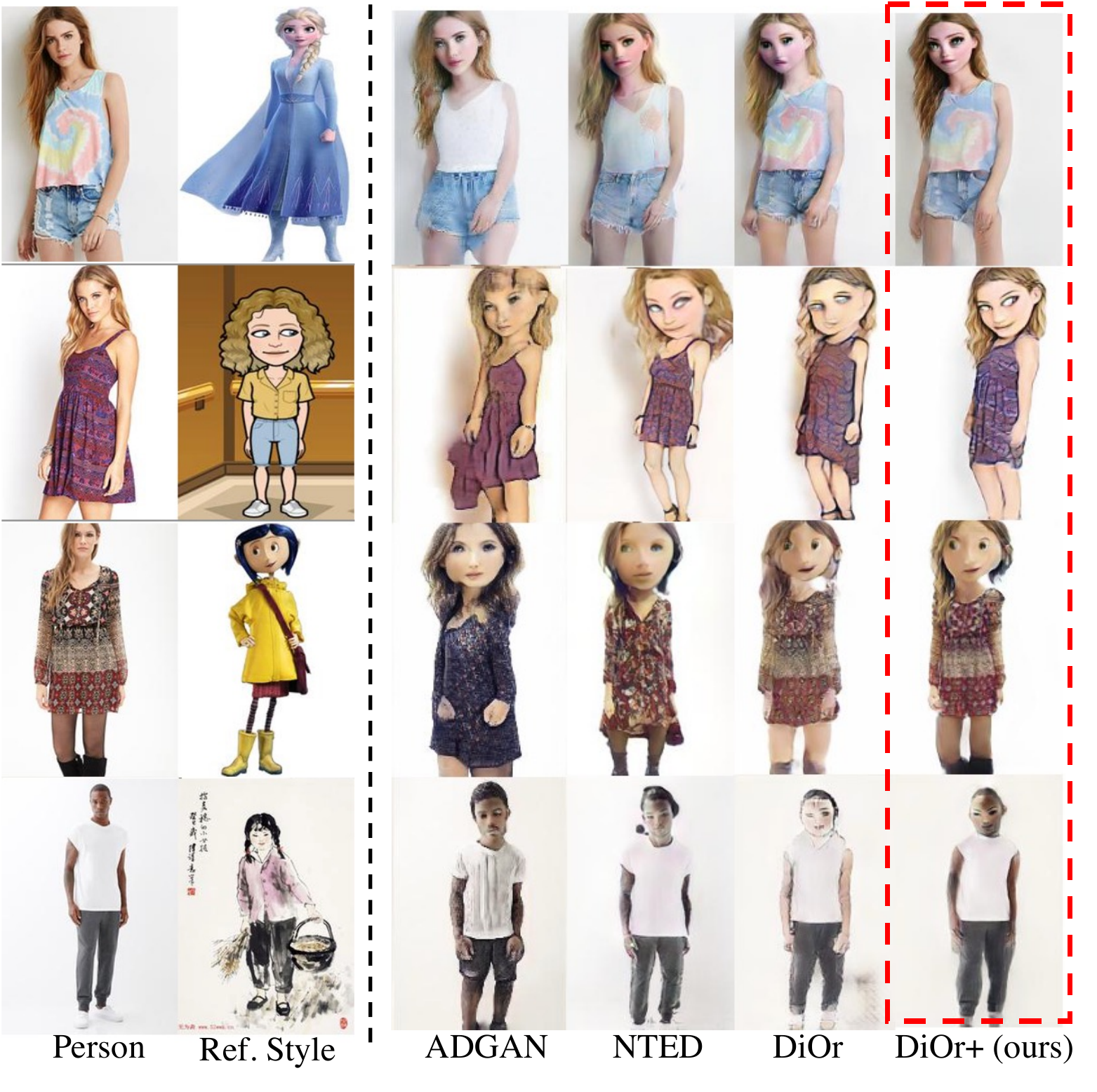}
   \vspace{-3mm}
  \caption{\textbf{Comparison for base model selections.} In each experiment, we apply the proposed stylization methods on different pose-guided person generators.}
   \vspace{-5mm}
  \label{fig:base_model_comparison}
\end{figure}

\begin{figure*}
  \centering
  \includegraphics[width=0.75
\textwidth]{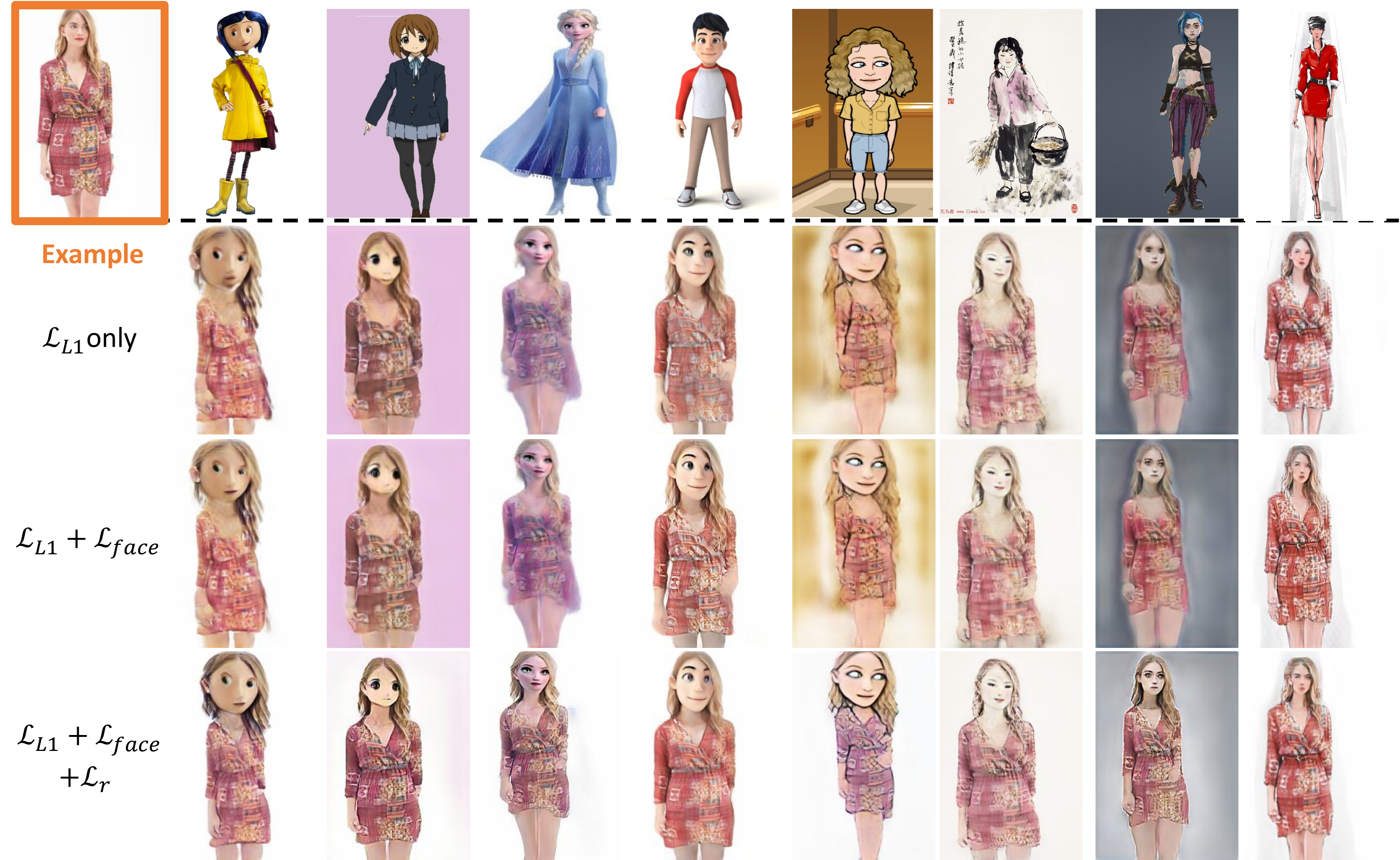}
  \vspace{-3mm}
  \caption{\textbf{Ablations Study} of the finetuning objectives. In the first two rows, we show the effectiveness of the L1 loss and face loss on restylization. The last row shows the effectiveness from adding a joint training of style transfer task on an auxiliary person images dataset.   }
  \label{fig:ablation}
  \vspace{-5mm}
\end{figure*}
\subsection{Comparisons with Universal Style Transfer}

Fig. \ref{fig:sota_comparison} shows a visual comparison of our results with state-of-the-art example-based universal style transfer methods: AdaConv~\cite{chandran2021adaconv}, NNST~\cite{kolkin2022nnst}, and DST~\cite{kim2020dst}. For NNST, we use segmentation masks as conditional inputs to guide the style transfer. For DST, we use the detected human keypoints for the human images and the labeled keypoints for the art reference as the required input pairs of anchor points. We can see that AdaConv and NNST can only stylize the image pixel-wise, so they are not suitable for this task requiring body deformation. DST~\cite{kim2020dst} can perform deformation to some extent, but with strong artifacts. Plus, all universal style transfer methods fail to stylize the face properly.

We ran a user study to compare our methods with NNST, which has the best qualitative results among the above methods. Given that NNST is unable to perform body deformation, for fairness, we modify the input skeleton to the desired body ratio and use our generator $\mathbf{G}_0$ to render the rescaled person before running NNST. As shown in Table \ref{tab:user_study}(a), our method is preferred to NNST in {76.1\%} of comparisons.

\subsection{Base Model Selection}
Although our proposed stylization method is designed to fit any end-to-end pose transfer model conditioned on keypoints, as discussed in Section 3, the selection of the base model has a significant impact on stylization quality. 
Fig. \ref{fig:base_model_comparison} gives a qualitative comparison of stylization performance with several base models: ADGAN~\cite{men2020adgan}, NTED~\cite{ren2022nted}, DiOr~\cite{Cui_2021_dior}, and our enhanced DiOr+. We downloaded the weights of all the above models from their official code repositories, and fine-tuned all models with the same procedure described in Section 3. As in Fig. \ref{fig:base_model_pose_transfer},  artifacts like broken faces or wrongly generated garments do not get fixed, but can only worsen, after the fine-tuning. ADGAN is insufficient for both outlier body ratios and outfit reconstruction. NTED is robust to extreme body ratios but suffers from reconstruction error. We further notice that it is harder to fine-tune NTED to get  outlier art faces correctly stylized because its decoder entangles the skin and the outfits; however, the former requires more overfitting to be stylized, while the latter requires less to preserve the content. These conflicts lead to difficulty in learning. DiOr, the flow-based method, is not always robust to extreme pose deformations which are underrepresented in the training set, especially for the face regions. 

In Tab. \ref{tab:model_selection}, we quantitatively compare base models using three proxy tasks: pose transfer and reconstruction on photo-realistic images, and reconstruction of art images by first destylizing them with $\mathbf{G}_0$ and then restylizing them with $\mathbf{G}_a$. We fine-tune all base models with the same objectives, but use a higher learning rate of $0.001$ for NTED because we find that it has a very minor response with the lower learning rate. 
For pose transfer and reconstruction, we report LPIPS \cite{zhang2018lpips} and FID \cite{heusel2017fid} metrics, and for art reconstruction, we report the average LPIPS among the {26} reconstructions only, because FID is not suitable when the {26} art images are not from the same domain.
We can see that NTED has the best performance for the transfer of realistic poses, though, as shown earlier in Fig. \ref{fig:base_model_pose_transfer}, it cannot reconstruct the outfit details. DiOr also has better pose transfer performance because it is predominantly trained on pose transfer tasks, while our DiOr+ is equally trained on pose transfer and inpainting tasks.  Our method is the best for image reconstruction as well as for art image reconstruction.

% NTED suffers from reconstructing the art characters because the reconstruction loss leads an inaccurate destylized art input, which will be further mistakenly recovered in the restylized step. DIOr suffers from recovering large faces and therefore performs worse than ours.

Additionally, we ran user studies to compare the stylization results of our DiOr+ model with those of NTED and the original DiOr. As shown in Tab.\ref{tab:user_study}(b), our model is preferred to NTED 70.6\% of the time and to DiOr 72.6\% of the time. We notice that users prefer NTED over our method when NTED can generate images with a better quality of hand/face while our method can better preserve the outfits.

\begin{table}[t]
\centering
\caption{Base Model Performance. We compute FID$\downarrow$~\cite{heusel2017fid} and LPIPS$\downarrow$~\cite{zhang2018lpips} in the photo domain for pose transfer and reconstruction task. We also compute LPIPS for art image reconstruction, which is first desylized and then stylized. We construct the ground truth distribution of FID from the test set of DeepFashion excluding images without faces. }
\vspace{-3mm}
\resizebox{\columnwidth}{!}{
\begin{tabular}{|l|ll|ll|l|}
\hline
& \multicolumn{2}{|c|}{Pose Trans.} & \multicolumn{2}{|c|}{Reconst.} & Art \\ \hline 
                
                & FID  & LPIPS  & FID  & LPIPS  & LPIPS \\ \hline
ADGAN \cite{men2020adgan} & 21.3 & 0.230 &  22.9 & 0.285  & 0.268 \\ 
%GFLA \cite{ren2020gfla} & - & - &  - & - & \\  %\hline
NTED \cite{ren2022nted} & \textbf{17.5} & \textbf{0.181} &  20.0 & 0.274 & 0.110  \\ 
DiOr \cite{Cui_2021_dior}& 18.8 & 0.190 &  17.0 & 0.190 & 0.098 \\ 
\hline
DiOr+ (ours)& 18.6 & 0.194 &  \textbf{14.6} & \textbf{0.129} & \textbf{0.059} \\ \hline
\end{tabular}
}
\label{tab:model_selection}
\vspace{-5mm}
\end{table}
\vspace{1ex}
\begin{table}[]
\centering
\caption{\textbf{User Study.} Group (a) is a comparison with the universal style transfer. In (b) we compare different base models. In (c), we compare with our ablation.  For each user study, we collect 340 responses from 45 users. }
\vspace{-3pt}
\begin{tabular}{|ll|l|}
\hline
                &  Comparisons & Preferred rates \\ \hline \hline
(a) & Ours vs. NNST \cite{kolkin2022nnst}  &   \textbf{76.1\%} vs. 23.9\%  \\ \hline 
\multirow{2}{*}{(b)} & Ours vs. NTED\cite{ren2022nted}  & \textbf{70.6\%} vs. 29.4\% \\ %\hline
& Ours vs. DiOr \cite{Cui_2021_dior}  & \textbf{72.6\%} vs. 27.4\% \\  \hline
(c) & Ours vs. version w/o $\mathcal{L}_{r}$  & \textbf{59.3\%} vs. 40.7\% \\ 
\hline
\end{tabular}
\label{tab:user_study}
\vspace{-7pt}
\end{table}

\begin{figure*}
  \centering
  \includegraphics[width=0.98\textwidth]{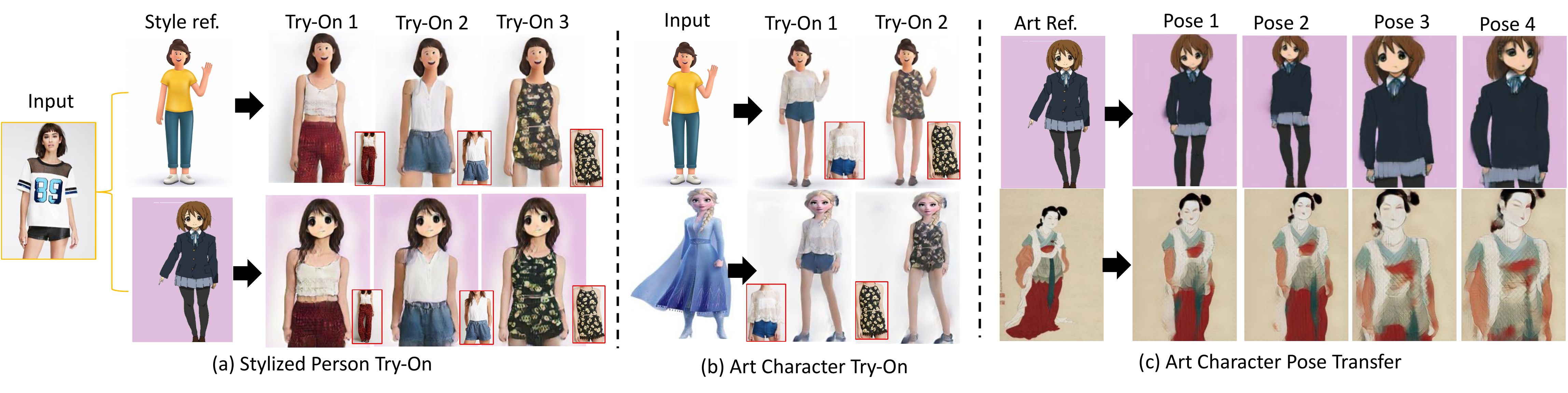}
  \caption{Applications.}
  \label{fig:extension}
  \vspace{-7mm}
\end{figure*}

\subsection{Ablation Study}
We verify our stylization objective by showing the effectiveness of our three loss terms, $\mathcal{L}_{L_1} + \mathcal{L}_{face}+ \mathcal{L}_{r}$. Fig.\ref{fig:ablation} shows qualitative comparisons. With only the art image $I_a$ for fine-tuning, the model is able to learn an overall stylization with the reconstruction loss $\mathcal{L}_{L_1}$, but face details are not confidently rendered. With the addition of the face loss $\mathcal{L}_{face}$, the face is more clearly rendered but the model may overfit to the style image, which causes misalignment and blurriness in garments. However, as explained in Section 3, if a dataset of real human photos is available, e.g., the DeepFashion dataset, one can jointly train with the stylization task on real human photo data using the $\mathcal{L}_{r}$ loss. The resulting model is less prone to overfitting the style reference, and gets the best quality of results, as shown by the bottom row of Fig.\ref{fig:ablation}.

Tab. \ref{tab:user_study}(c) shows the results of a user study comparing our full method with the ablated version lacking $\mathcal{L}_{r}$, i.e., the version that solely takes a single art image as the training data. Our full model is preferred 59.3\% of the time over the ablated version. While the full model still gives the best results on the whole, the smaller preference margin suggests that even without a regularization on photographic images, the proposed method can still give reasonable results.

\section{Discussion}
%This work proposes a generic method to stylize full-body human images by fine-tuning a pre-trained pose-guided person generation model. We address the unique challenges in this domain including deformation and stylization to stylize a human image fully.

\noindent \textbf{Extensions and applications.} As shown in Fig. \ref{fig:extension}, by manipulating the conditioning information given to the fine-tuned generator $\mathbf{G}_a$, we can inherit the use cases of the base human generation model, such as virtual try-on and pose transfer for art characters. Note, however, that the quality of pose transfer results is not very high because, as explained in Section 3.1, the training of our DiOr+ generator is more heavily weighted towards reconstruction rather than pose transfer, as well as because of the obvious limitations of single-shot fine-tuning on out-of-domain data.

\begin{figure}
  \centering
  \includegraphics[width=0.45\textwidth]{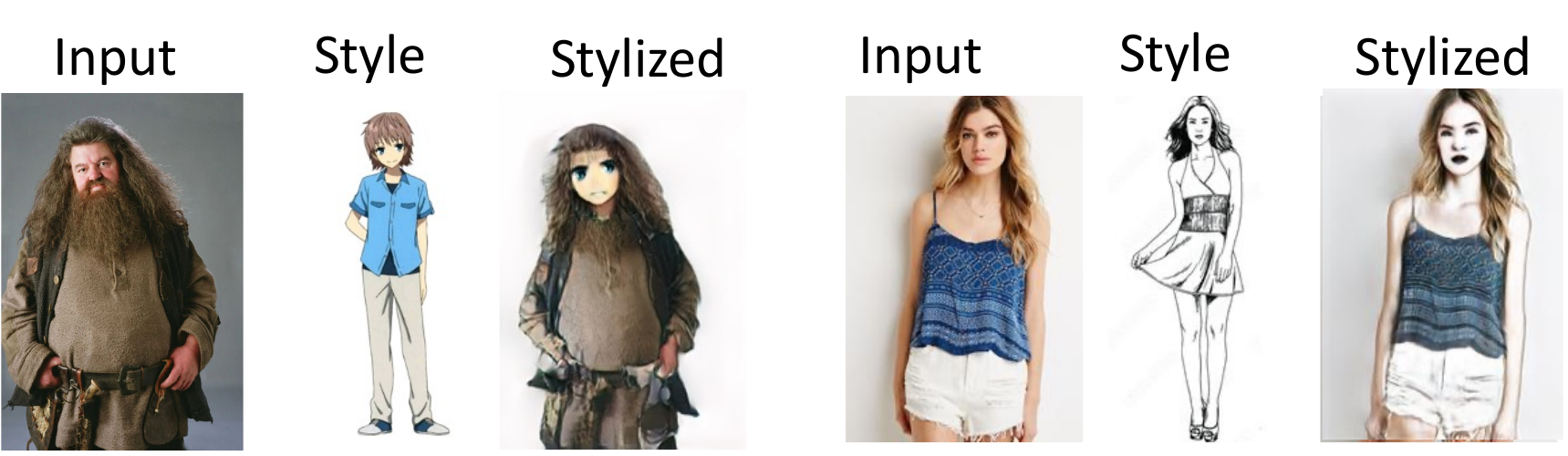}
  \caption{Failure cases.}
  \label{fig:failure}
  \vspace{-7mm}
\end{figure}

\noindent \textbf{Failure cases} are shown in Fig.\ref{fig:failure}. For the out-of-dataset input in the left example, our method fails to render the beard, which is an unseen facial element in the training. In the right example, the black-and-white style is not correctly learned -- the garment and arms are still in color -- because the destylization step is not sufficient to colorize the grayscale garment, so the fine-tuning has no opportunity to learn decolorization. 

\noindent \textbf{Limitations and future work.} Although we have achieved promising results, our output quality is largely dependent on the base generator for the realism, resolution, and warping quality. In the future, if stronger pre-trained human models become available, our fine-tuning approach should be able to achieve better results.

\paragraph{Acknowledgements.} We thank Viraj Shah for very helpful discussions and
feedback.

%In addition, it is hard and subjective to define how much the garments should be stylized to both reflect the given style and maintain their original texture. In this work, we decide to maintain the garment as much as the original, though some human raters may consider it as not stylized enough. In the future, we plan to design a controllable way to determine the stylization level of garments. 
%%%%%%%%% REFERENCES

{\small
\bibliographystyle{ieee_fullname}
\bibliography{egbib}
}

\onecolumn

%\addappheadtotoc
\appendix
\appendixpage
\startcontents[sections]
\printcontents[sections]{l}{1}{\setcounter{tocdepth}{2}}

\section{User Study Interface}
Fig. \ref{supp_fig:user_study} shows a screenshot of the user study interface used in this project.

\begin{minipage}[c]{\linewidth}

    \makebox[\linewidth]{
        \includegraphics[width=0.7\textwidth]{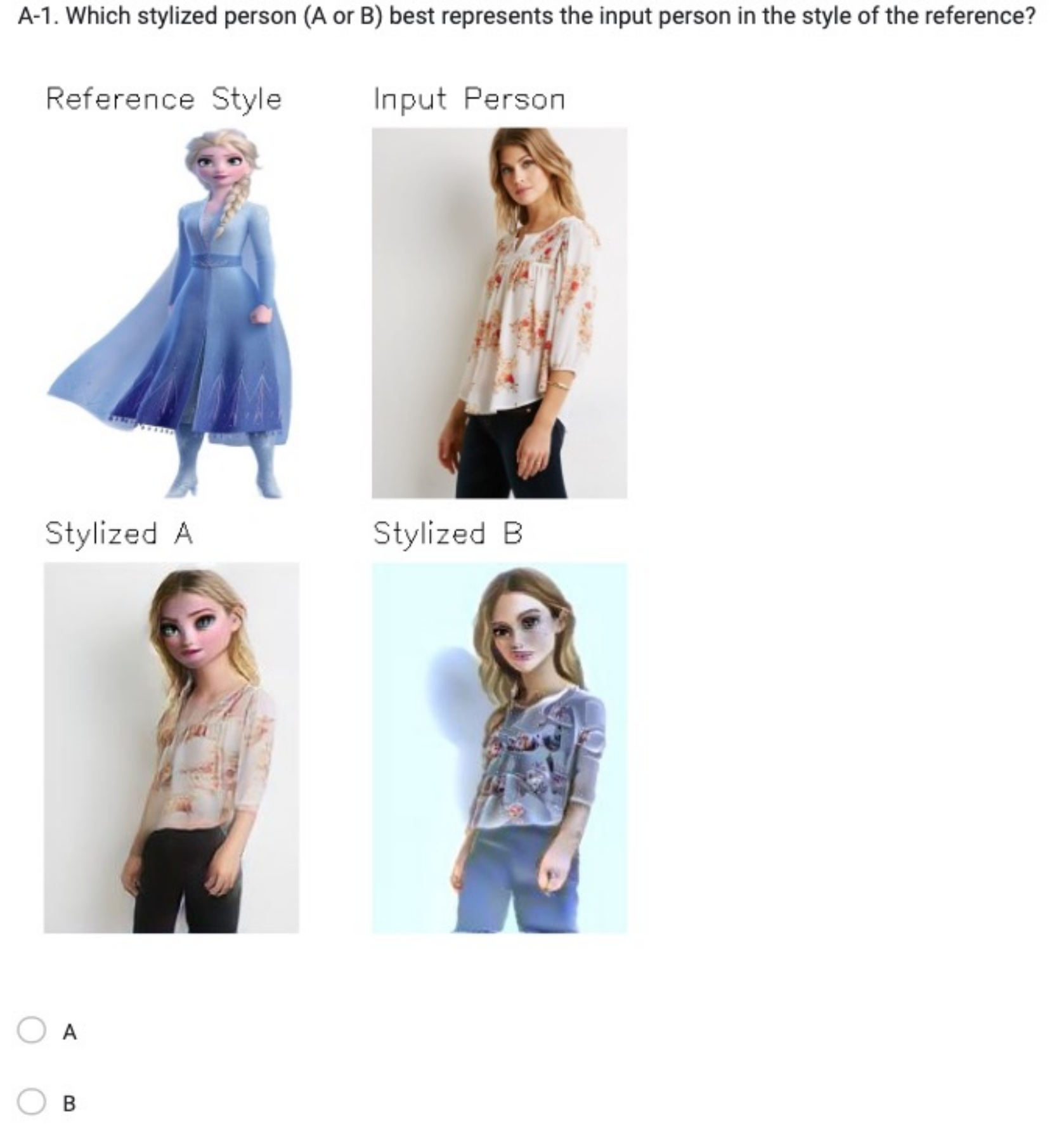}
    }
    \captionof{figure}{\textbf{User Study Interface.}}
    \label{supp_fig:user_study}
\end{minipage}
\newpage

\section{Destylization and Restylization Details}
Here we show the destylization and restylization with different base models, NTED~\cite{ren2022nted}, DiOr~\cite{Cui_2021_dior}, and DiOr+ (ours). 
NTED is an encoder-decoder-based model which generates the images from conditional embeddings rather than explicitly warps the spatial contents like the other two methods. Consequently, NTED is not able to reconstruct the out-of-distribution art images in the destylization steps, which causes learning difficulty in the restylization step to reconstruct the art image. 
DiOr is a warping-based method, which takes an appearance input and uses the mean background vector to broadcast the background regions to generate the background while warping the garments in the foreground and using the mean skin vector to generate the skin/face areas. The other two methods do not have regional conditioning for the background generation, so NTED and DiOr+ are easier to get overfitting in the stylization steps. Moreover, since DiOr is a warping method and not seeing large faces during training, the large face (e.g. the second art example) can't get fully reconstructed.

\begin{minipage}[c]{\linewidth}
    \centering
    \makebox[\linewidth]{
        \includegraphics[width=\linewidth]{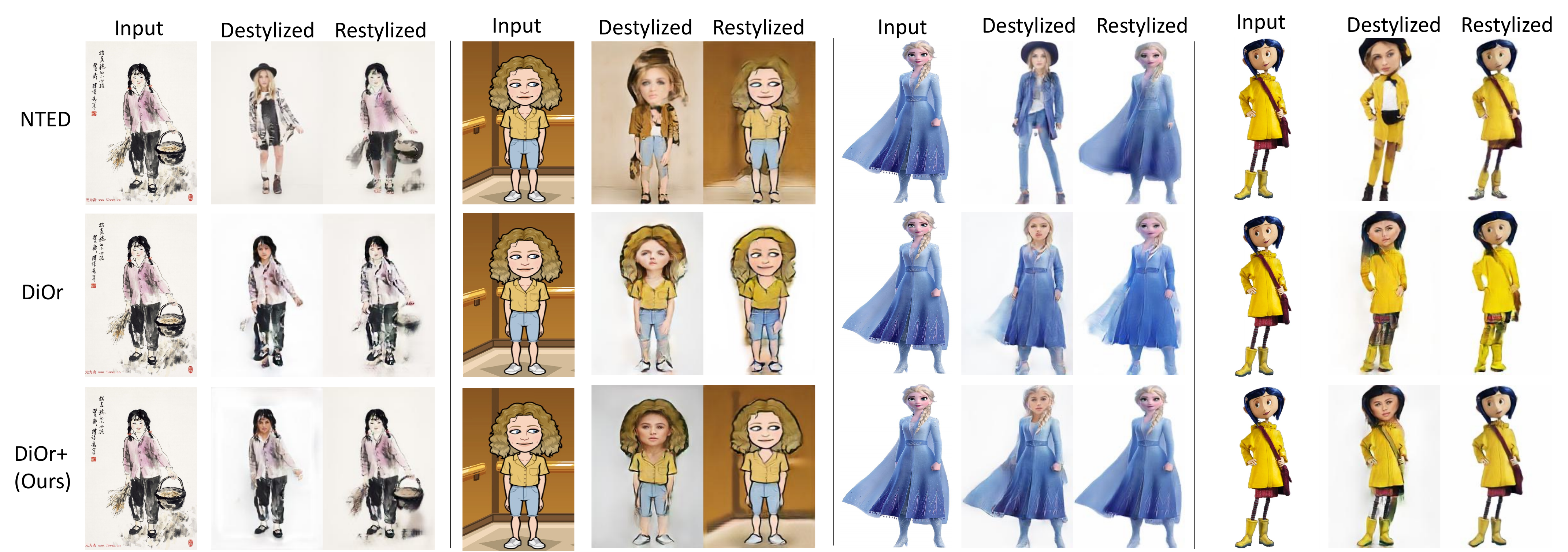}
    }
    \captionof{figure}{\textbf{Destylization and Restylization} of different methods on art images.}
    \label{supp_fig:destylization}
 \end{minipage}
\newpage
\section{Skeleton Deformation Details}
In Fig. \ref{supp_fig:deformation}, we show the skeleton deformation effects. We first show that without skeleton deformation, only pixel-wise stylization of a person image is insufficient to fully represent the style. 

Next, we show a naive baseline skeleton deformation method. Given a reference skeleton $\Phi_a = \{(\alpha_i^a,l_i^a)\}$ and a target skeleton $\Phi_p = \{(\alpha_i^p,l_i^p)\}$, where $i=0,...,17$, we rescale the skeleton $\Phi_p$ towards the reference body ratio by keeping the angles and swapping the lengths, so the directly scaled skeleton is $\Phi''_p = \{(\alpha_i^p,l_i^a)\}$. From the results in Fig. \ref{supp_fig:deformation}, we observe that the projected body segment lengths in 2D cannot be directly swapped when two skeletons have very different orientations; otherwise, the scaled skeleton would be distorted and incorrect.

Finally, with our proposed learnable skeleton deformation module, the skeleton can be plausibly scaled and gives reasonable outputs.

\begin{minipage}[c]{\linewidth}
    \centering
    \makebox[\linewidth]{
        \includegraphics[width=\linewidth]{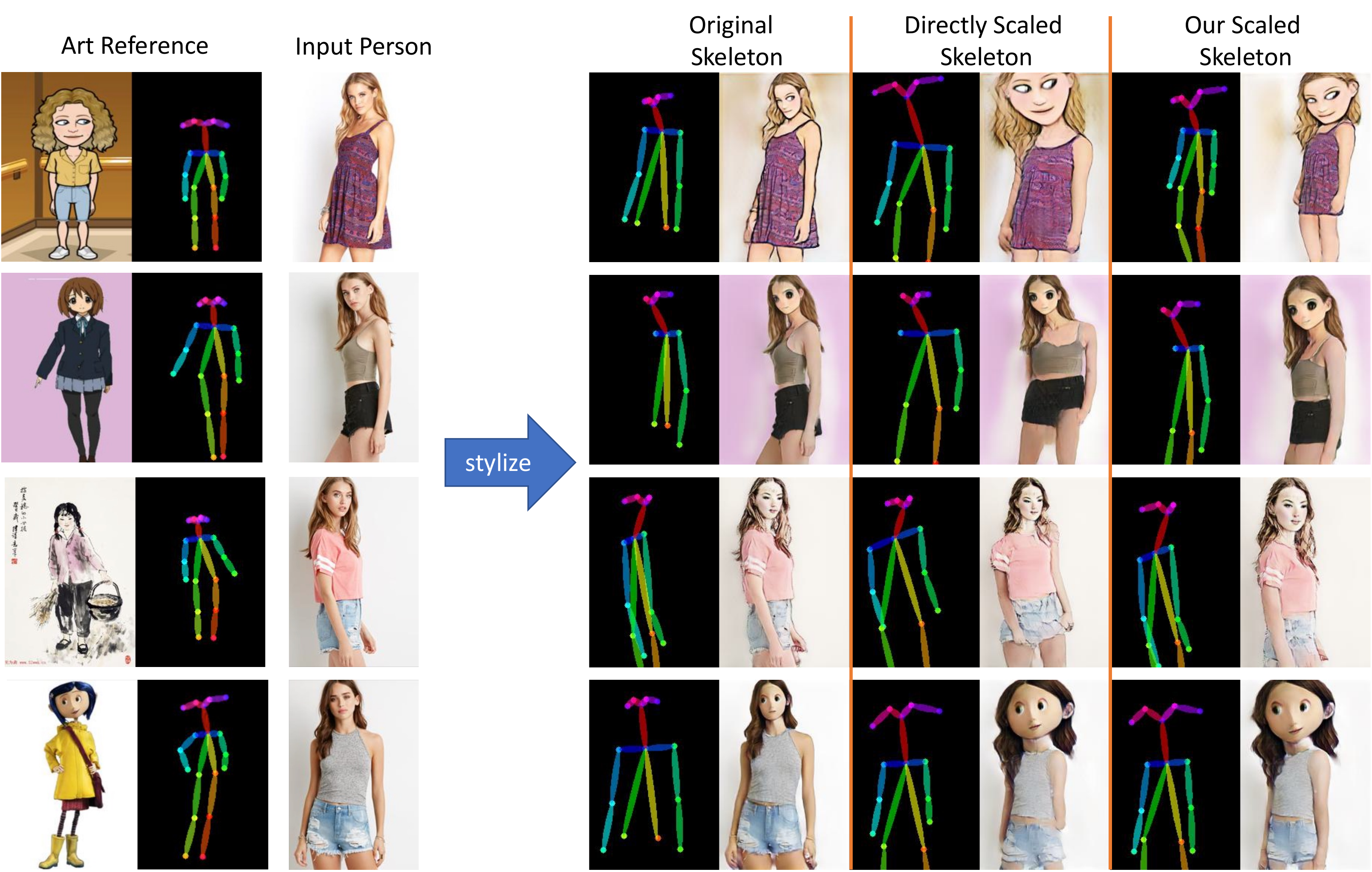}
    }
    \captionof{figure}{\textbf{Skeleton Deformation Comparisons.} We show the stylization results conditional on the original skeletons, directly scaled skeletons and the skeleton scaled by our proposed learnable deformation module. The art input and generation outputs are shown together with the skeleton visualization.}
    \label{supp_fig:deformation}
 \end{minipage}
 
 \newpage
 
\section{More Examples for Stylization}

We show more results for our stylization. In most cases, our method can stylize the given input towards the art character plausibly. However, we notice that the fine-tuning inherits the limitations from the base model, like misaligned warping, inconsistent skin colors for limbs, or missing hands, which are common problems for full-body human generators trained on DeepFashion. This suggests that once stronger full-body human generators become available, our fine-tuning pipeline will yield better stylization results.

\begin{minipage}[c]{\linewidth}

    \makebox[\linewidth]{
        \includegraphics[width=0.9\textwidth]{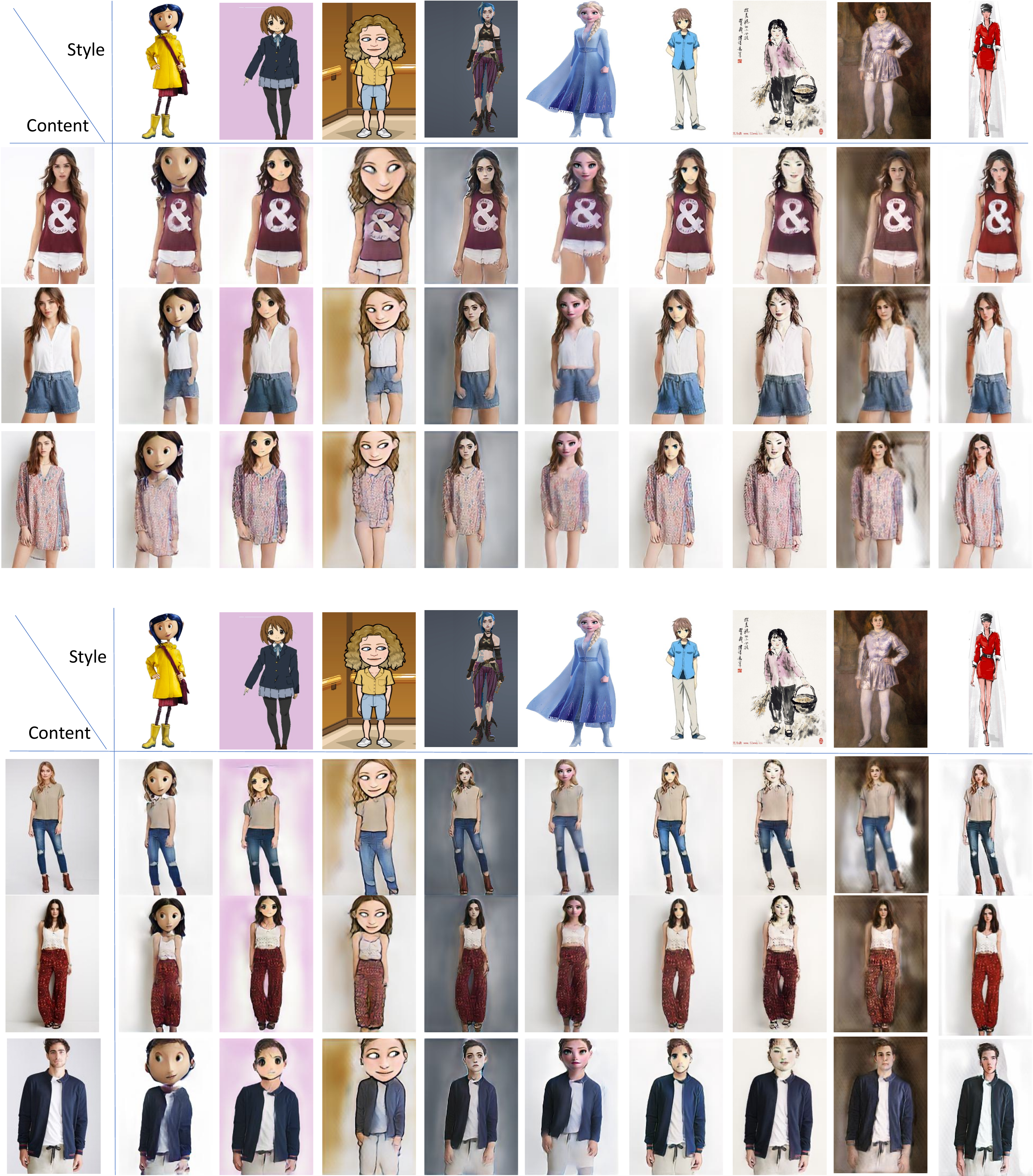}
    }
    \captionof{figure}{\textbf{More examples of our full stylization pipeline.}}
    \label{supp_fig:stylization}
 \end{minipage}
 \newpage
 
\section{More Examples for Comparisons with Generic Style Transfer Methods}
Here we show comparisons with the generic style transfer methods AdaConv \cite{chandran2021adaconv}, DST \cite{kim2020dst} and NNST \cite{kolkin2022nnst}. 
Note only DST can perform deformation with a set of keypoints for guidance as inputs. For NNST, we also show results using pre-processed person images as input, which are first rescaled by our learnable deformation module in the second last column in each group in Fig. \ref{supp_fig:supp_styletransfer}. From Fig. \ref{supp_fig:supp_styletransfer}, we can see our methods can transfer the style in a more faithful way than the others. 

\begin{minipage}[c]{\linewidth}

    \makebox[\linewidth]{
        \includegraphics[width=1\textwidth]{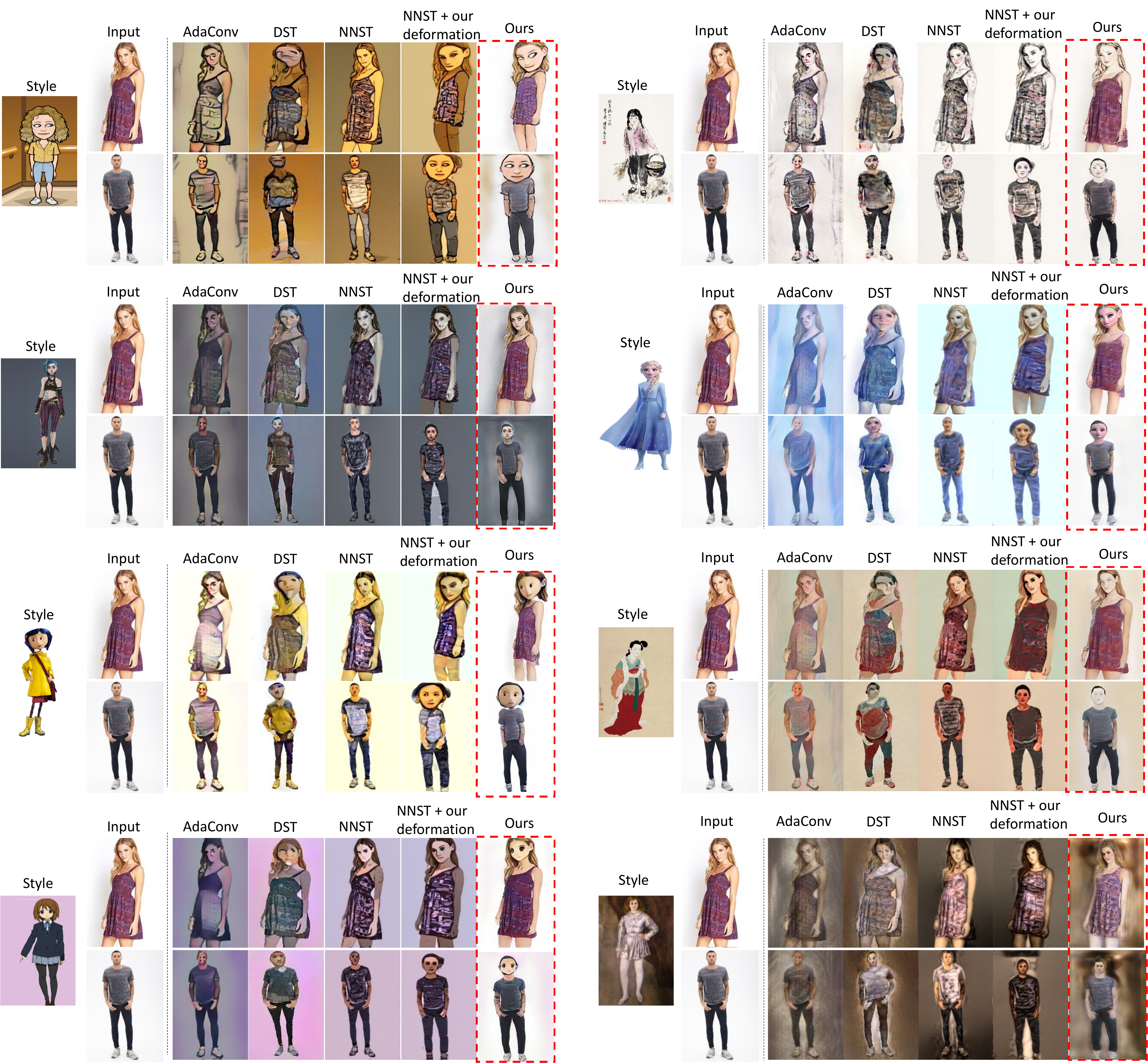}
    }
    \captionof{figure}{\textbf{Comparisons to Generic Style Transfer Methods.}}
    \label{supp_fig:supp_styletransfer}
\end{minipage}
\newpage

\section{More Examples for Comparisons with Different Base Models}
Here we show more examples of comparisons among different base model choices: ADGAN \cite{men2020adgan}, NTED \cite{ren2022nted}, DiOr \cite{Cui_2021_dior}, and DiOr+ (ours). All base models are fine-tuned by our method in the same way.

\begin{minipage}[c]{\linewidth}

    \makebox[\linewidth]{
        \includegraphics[width=1\textwidth]{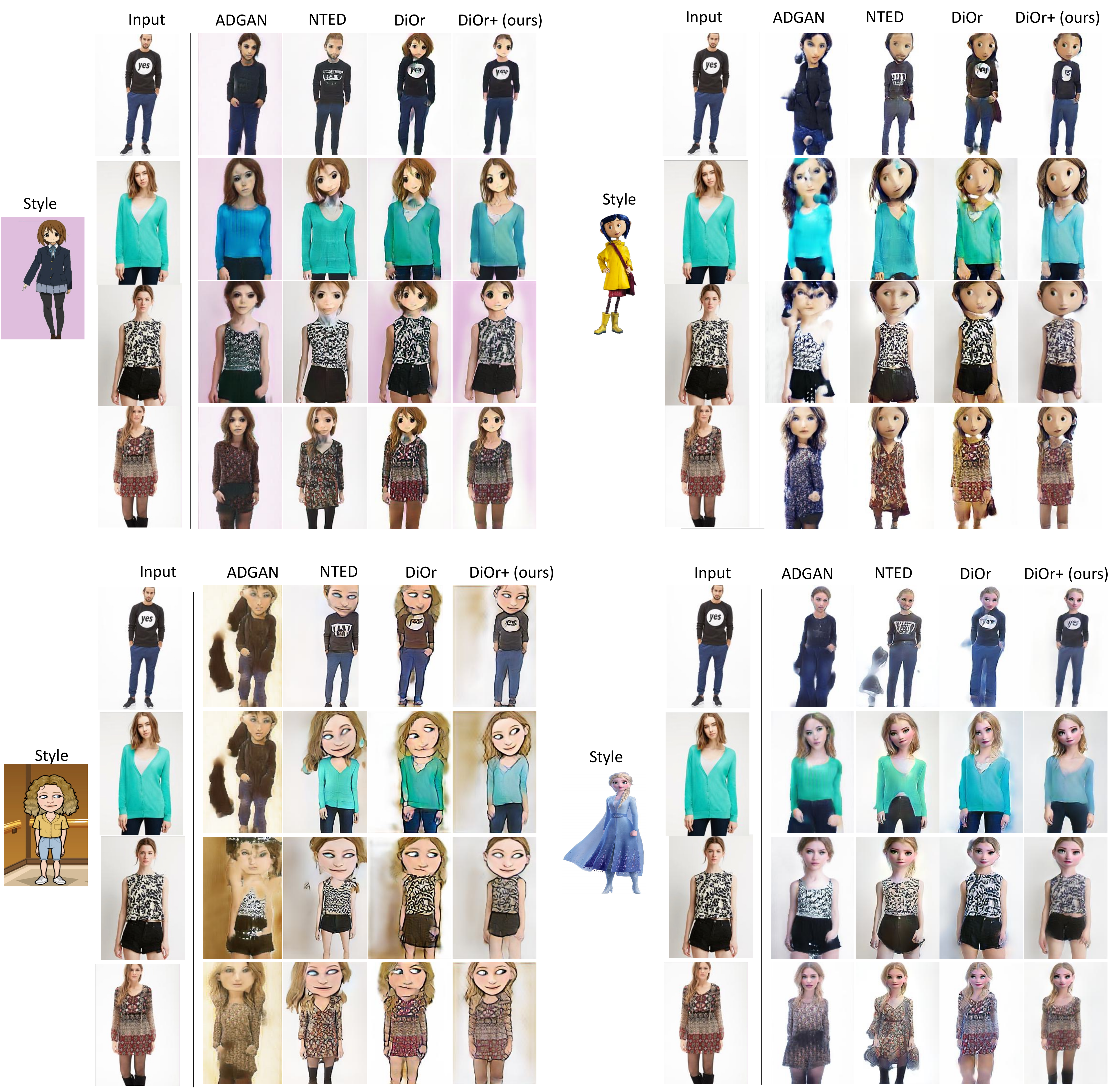}
    }
    \captionof{figure}{\textbf{Comparisons to Generic Style Transfer Methods.}}
    \label{supp_fig:base_model}
\end{minipage}

\newpage
\section{More Examples for Ablation Study }

\begin{minipage}[c]{\linewidth}

    \makebox[\linewidth]{
        \includegraphics[width=0.77\textwidth]{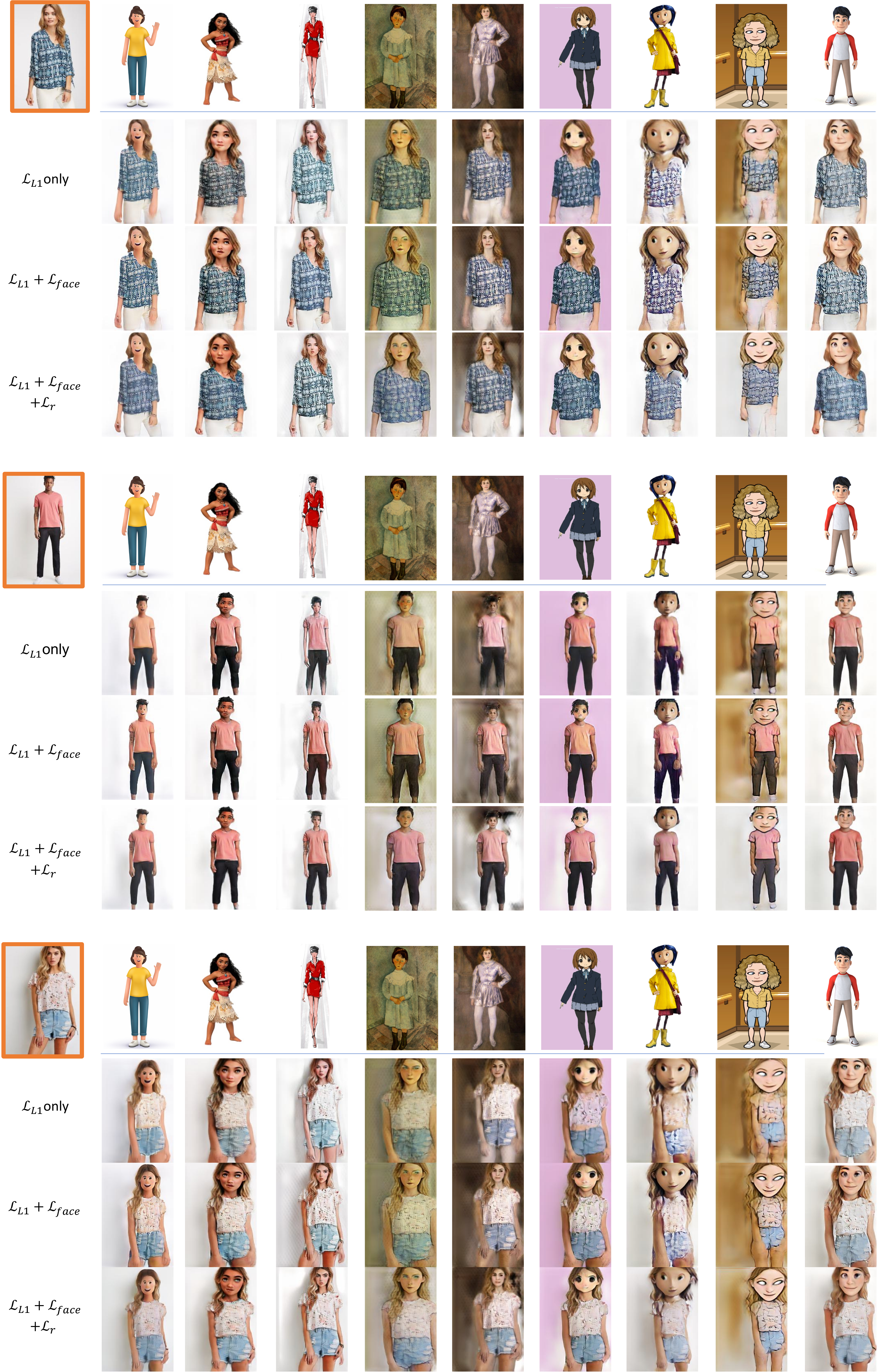}
    }
    \captionof{figure}{\textbf{Ablation Study} of the stylization objectives.}
    \label{supp_fig:ablation}
\end{minipage}

\section{Art Image Gallery}

\begin{minipage}[c]{\linewidth}

    \makebox[\linewidth]{
        \includegraphics[width=0.8\textwidth]{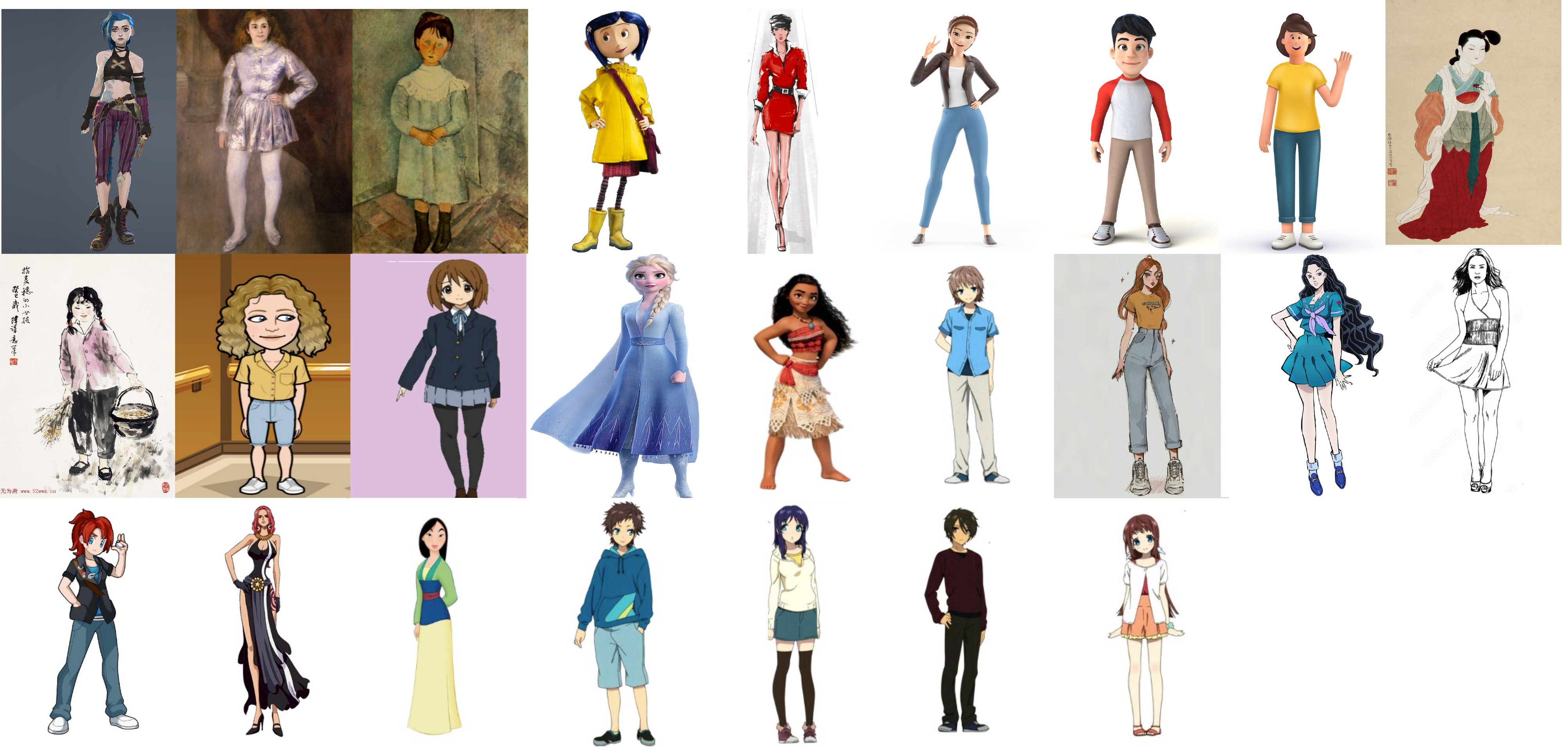}
    }
    \captionof{figure}{\textbf{Gallery of Art Images.}}
    \label{supp_fig:supp_gallery}
\end{minipage}
Here we show the gallery of the 25 art images collected from the Internet and used in this project in Fig. \ref{supp_fig:supp_gallery}.
The images are collected from the below sources:
\begin{itemize}
    \item  Anime, \textit{Nagi no Asukara} from 	P.A. Works
    \item Anime, \textit{K-On!} from Kyoto Animation
    \item Anime, \textit{JoJo's Bizarre Adventure} from 	Shueisha 
    \item Anime, \textit{One Piece} from Shueisha
    \item TV series, \textit{Arcane} from Netflix
    \item Film, \textit{Coraline} from Laika
    \item Film, \textit{Frozen} from Walt Disney Animation Studios
    \item Film, \textit{Moana} from Walt Disney Animation Studios
    \item Film, \textit{Mulan} from Walt Disney Feature Animation Studios
    \item Painting, \textit{Madame Heriot en travesti} by Pierre-Auguste Renoir
    \item Painting, \textit{Little Girl In Blue} by Amedeo Modigliani
    \item Painting, \textit{Little Girl Gleaning} by Mou Chen, \\\url{http://www.sinaimg.cn/dy/slidenews/26_img/2014_20/18496_348224_651181.jpg}
    \item Painting, \textit{lady portrait} by Zhenzhi Jin, \\ \url{https://images.twgreatdaily.com/images/image/pfi/pfivnnAB3uTiws8KVWq9.jpg}
    
    \item Bitimoji, \url{bitmoji.com}
    \item DeviantArt, \url{https://www.deviantart.com/ravenide/art/Pokemon-Trainer-Alex-Fullbody-608209332}
    
    \item Shutterstock with standard licenses, \url{https://www.shutterstock.com/}
    \item Adobe Stock with standard licenses, \url{https://stock.adobe.com/}
    \item Pinterest, \url{https://www.pinterest.com/}
\end{itemize}

\end{document}